\documentclass[sigconf,authorversion,nonacm]{acmart}
\AtBeginDocument{%
  }

\setcopyright{acmlicensed}
\copyrightyear{2025}
\acmYear{2025}
\acmDOI{XXXXXXX.XXXXXXX}
\acmConference[KDD '25]{Proceedings of the 31st ACM SIGKDD Conference on Knowledge Discovery and Data Mining V.1}{August 3--7, 2025}{Toronto, ON, Canada}
\acmISBN{978-1-4503-XXXX-X/2018/06}

\begin{document}

\title{Multimodal Banking Dataset: Understanding Client Needs through Event Sequences}


\author{Dzhambulat Mollaev}
\affiliation{
  \institution{Sber AI Lab}
  \city{Moscow}
  \country{Russia}
}
\email{dzhambulat.mollaev2012@yandex.ru}
\author{Ivan Kireev}
\author{Mikhail Orlov}
\author{Alexander Kostin}
\author{Ivan Karpukhin}
\affiliation{
  \institution{Sber AI Lab}
  \city{Moscow}
  \country{Russia}
}
\author{Maria Postnova}
\affiliation{
  \institution{Sber}
  \city{Moscow}
  \country{Russia}
}
\author{Gleb Gusev \orcid{0009-0003-7298-1848}}
\affiliation{
  \institution{Sber AI Lab}
  \city{Moscow}
  \country{Russia}
}
\author{Andrey Savchenko \orcid{0000-0001-6196-0564}}
\affiliation{
  \institution{Sber AI Lab}
  \city{Moscow}
  \country{Russia}
}
\email{avsavchenko@hse.ru}

\renewcommand{\shortauthors}{Mollaev et al.}

\begin{abstract}
  Financial organizations collect a huge amount of temporal (sequential) data about clients, which is typically collected from multiple sources (modalities). Despite the urgent practical need, developing deep learning techniques suitable to handle such data is limited by the absence of large open-source multi-source real-world datasets of event sequences. To fill this gap, which is mainly caused by security reasons, we present the first industrial-scale publicly available multimodal banking dataset, MBD, that contains information on more than 2M corporate clients of a large bank. Clients are represented by several data sources: 950M bank transactions, 1B geo position events, 5M embeddings of dialogues with technical support, and monthly aggregated purchases of four bank products. All entries are properly anonymized from real proprietary bank data, and the experiments confirm that our anonymization still saves all significant information for introduced downstream tasks. Moreover, we introduce a novel multimodal benchmark suggesting several important practical tasks, such as future purchase prediction and modality matching. The benchmark incorporates our MBD and two public financial datasets. We provide numerical results for the state-of-the-art event sequence modeling techniques including large language models and demonstrate the superiority of fusion baselines over single-modal techniques for each task. 
  Thus, MBD provides a valuable resource for future research in financial applications of multimodal event sequence analysis.
  
HuggingFace Link: \url{https://huggingface.co/datasets/ai-lab/MBD}
 
 Github Link: \url{https://github.com/Dzhambo/MBD}
\end{abstract}



\received{10 February 2025}

\maketitle

\section{Introduction}\label{sec:intro}
In the key tasks in the banking industry, such as campaigning, fraud detection, credit risk assessment, customer segmentation, and personalized recommendations, the performance heavily depends on the processing of the client's financial and non-financial activities, e.g., transactions, dialogues with sales managers, product purchase history, etc.~\cite{babaev2022coles,10.1145/502512.502581}. This data, spanning extended periods, is typically annotated with temporal information, forming what is known as {\it event sequences}~\cite{Udovichenko_2024, yeshchenko2022surveyapproacheseventsequence, system_call_sequences,  Rare_Events, guo2020surveyvisualanalysisevent}. An event is described by several heterogeneous fields, numerical and categorical. An essential property of event sequences is that these data are often gathered from multiple sources or channels, rendering them {\it multimodal}. 

The success of financial organizations strongly depends on their availability to deeply analyze such multi-source heterogeneous event sequences for building classification and predictive models. 
However, existing multimodal models~\cite{xu2023multimodallearningtransformerssurvey,zhang2023crossformer} cannot be directly applied to such event/tabular data due to their significant difference from audio, images, texts, and regular time series. Unfortunately, despite the urgent business needs, the progress in the development of multimodal techniques for multi-source event sequences is limited by the absence of large-scale datasets. Indeed, though several datasets of event sequences are used in research, e.g., credit card transactions~\cite{padhi2021tabular} or MIMIC~\cite{mimic4}, they are either small or contain only one modality. Thus, tackling the complexity of multimodal event sequence data is still very challenging.

To bridge this gap caused mainly by security reasons in the fintech industry, this paper introduces the Multimodal Banking Dataset (MBD), an unprecedented open-source resource encompassing extensive multichannel event sequence data of banking corporate clients. It is the largest dataset of its kind, featuring detailed records of approximately 2 million clients across four distinct modalities: money transfers (about 950 million events), geo position data (around 1 billion events), technical support dialog embeddings (approximately 5 million entries), and monthly aggregated bank product purchases categorized into four types. Each modality encompasses roughly one or two years of historical, time-annotated data, making it a rich resource for analyzing the dynamics of client behavior over time.

The MBD dataset enables the research of several critical business problems in a multimodal context, such as future purchase prediction (campaigning) and matching different modalities for the same clients. To support these tasks, we provide benchmarks using the MBD and other existing financial datasets of much smaller size. 
To ensure client privacy, all data in MBD has been rigorously anonymized, allowing researchers to work with the dataset without compromising confidentiality. Moreover, our experiments confirm that our anonymization procedure preserves the consistency of model performance between original and anonymized data.

\section{Related Works}\label{sec:rel}

\subsection{Financial data}
Banks' multitude of services and processes generates a variety of data that can be considered as modalities. Early works \cite{bank_telemarketing, mancisidor2021learning} use feature processing techniques that remove multimodal complexity and present data in tabular form. A wide range of financial aggregates are presented in the American Express dataset \cite{amex} as a sequence of historical slices. 
The development of deep learning methods has led to the ability to work with complexly structured data, such as sequences of events. Quite a few datasets \cite{agepred, rosbank, alphabattle}, mostly unimodal, presented mainly at ML competitions. To work with such data, both supervised \cite{ala2022deep, babaev2019rnn, wang2022deep} and unsupervised methods \cite{padhi2021tabular, bazarova2024universal, Skalski2023TowardsAF} are used. 
A multimodal financial sequential dataset was introduced in DataFusion 2022 competition~\cite{datafusion2022}. There are two sequential modalities, transaction and web clickstream, and two downstream tasks: matching and education level prediction. However, this is an extremely small dataset of 22K clients, and no accurate baseline model is available.

\subsection{Other event sequence domains}
Temporal point process \cite{mei2017neural, Zhuzhel2023ContinuoustimeCM} model streams of discrete events in continuous time by constructing a neurally multivariate point process. The authors use a large collection of single-modal datasets from different types of modalities: Media (Retweets \cite{zhao2015seismic}, MemeTrack \cite{leskovec2014snap}, Amazon \cite{amazon}, IPTV \cite{luo2014you}), Medical (MIMIC-II \cite{johnson2016mimic}), Social (Stack Overflow \cite{leskovec2014snap}, Linkedin \cite{xu2017dirichlet}) and Financial (Transaction \cite{fursov2021gradient}) data. Multimodal medical record datasets, MIMIC-IV \cite{mimic4} is used in
EventStreamGPT~\cite{mcdermott2024eventstreamgpt}. The authors propose a GPT-like approach for continuous-time event sequences. This dataset's structure is close to our MDB. However, unlike medical information, financial data contains longer chains of events and more regular patterns, and individual transactions are less informative. 

\subsection{Geostream and dialogues}
Geodata is used for various tasks. One of the uses of geo is a visualization of analytics on a map~\cite{Hao2011VisualSA}. However, geo is used here not as a separate modality but as additional tags to the mainstream of tweets. Mobile marketers \cite{Baye2024CustomerRA} use geo-targeting for pricing and send personalized recommendations. In~\cite{Verma2020GeoHashTB}, geo hashes are used for user mobility detection and prediction.

We encode our dialogue entries via a pretrained NLP model. Such models have already been used \cite{Hassan2019AutomaticAO} for text anonymization tasks. Embeddings preserve the meaning of the text, which was shown in \cite{Vaswani2017AttentionIA}.  Pre-trained text embeddings can capture text sentiment and improve text-to-speech models \cite{Hayashi2019PreTrainedTE}.

\section{Proposed Dataset}\label{sec:dataset}
\subsection{Overview of the MBD dataset}\label{subsec:mbd_overview}
\begin{figure*}[h]
  \centering
  \hspace*{-0.1\textwidth}\includegraphics[width=1.1\textwidth]{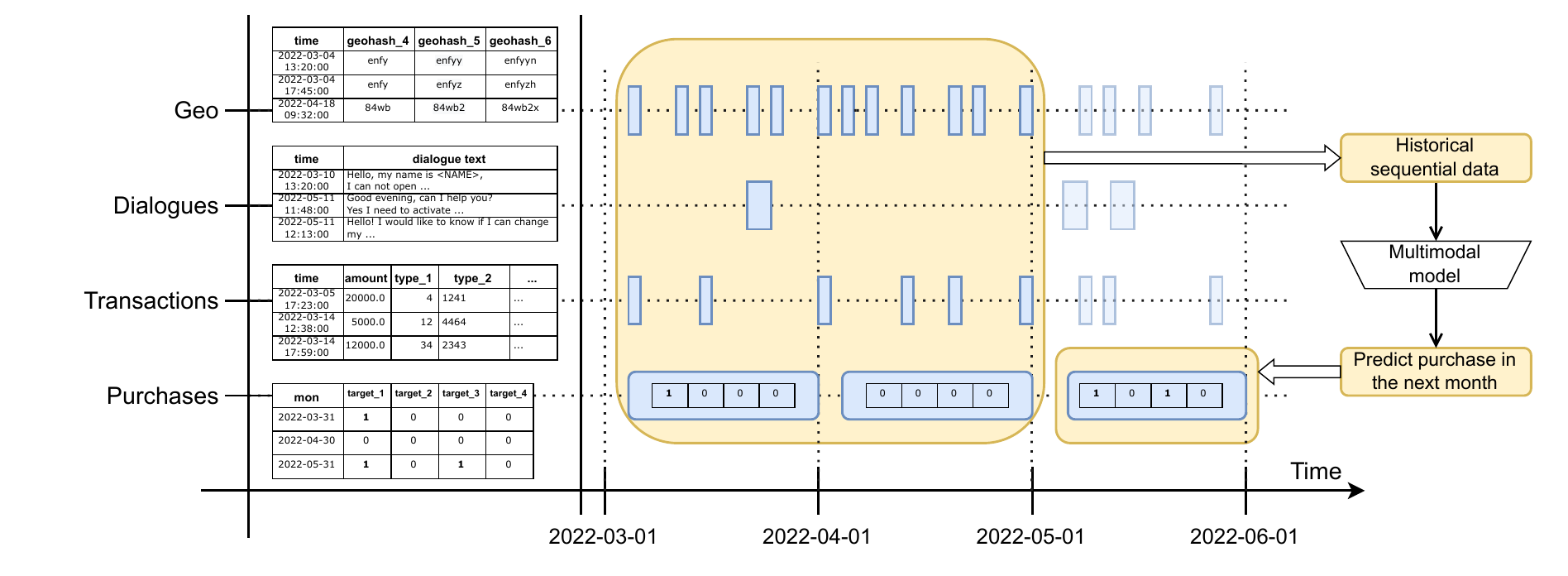}
  \caption{Original data sources processing pipeline for solving campaigning task. Examples of raw data for each modality are presented on the left. Only anonymized data is published in the dataset. The center shows the temporal anonymized structure of the data. The multimodal multi-label classification model for predicting purchases is shown on the right.}
  \Description{Pipeline for processing original data sources for solving campaigning task.}
 \label{fig:schema-downstream}
\end{figure*}

\begin{table*}
\caption{Overview of existing transaction datasets.}
\label{table:datasets_stat}
\resizebox{\textwidth}{!}{
\begin{tabular}{cccccc}
\toprule
\textbf{Dataset} & \textbf{\# Clients} & \textbf{Downstream Tasks} & \textbf{\# Events} & \textbf{Class Balance} & \textbf{Modalities} \\
\midrule
\textbf{Datafusion}~\cite{datafusion2022} & 22K & \begin{tabular}[c]{@{}c@{}}Binary classification\\ Multimodal matching\end{tabular} & 146M & Imbalanced & Transactions, Clickstream \\
\textbf{Alphabattle}~\cite{alphabattle} & 1.5M & Binary classification & 443M & Imbalanced & Transactions \\
\textbf{Age}~\cite{agepred} & 50K & Multiclass classification & 44M & Balanced & Transactions \\
\textbf{Rosbank}~\cite{rosbank} & 10K & Binary classification & 1M & Imbalanced & Transactions \\
\textbf{Credit Card Transaction}~\cite{padhi2021tabular} & 2K & \begin{tabular}[c]{@{}c@{}}Binary classification\\ Regression task\end{tabular} & 2M & Highly Imbalanced & Transactions \\
\textbf{MBD-small (ours)} & 70K & \begin{tabular}[c]{@{}c@{}}Multilabel binary classification\\ Multimodal matching\end{tabular} & 80M &  Imbalanced & \begin{tabular}[c]{@{}c@{}}Transactions, Geostream,\\ Dialogues\end{tabular} \\
\textbf{MBD (ours)} & 2M & \begin{tabular}[c]{@{}c@{}}Multilabel binary classification\\ Multimodal matching\end{tabular} & 2B & Highly Imbalanced & \begin{tabular}[c]{@{}c@{}}Transactions, Geostream,\\ Dialogues\end{tabular} \\
\bottomrule
\end{tabular}
}
\end{table*}

Modern innovative banking institutions actively develop AI technologies to customize their human-oriented technologies and make everyday decisions. A superior level of technology will lead to new customer experience cases, which should form a competitive advantage for the services provided, including the speed, accuracy, and price of customer services, including personal credit conditions, individual finance strategy, etc.

One of the main benefits of using AI is the ability to analyze large amounts of customer data. This helps banks better understand their customers' needs and offer them the most suitable products and services. Additionally, AI can protect customers from fraud and prevent financial losses. More accurate forecasting of financial risks associated with lending or investing allows us to provide more favorable conditions to more reliable clients. This strategy is more effective if there is a lot of data, so banks strive to accumulate as much information as possible. 

Being a team of a bank that stores petabytes of data about bank processes and clients, we understand the urgent needs of fintech data scientists in large sets of publicly available temporal data from various sources (bank transactions, client locations, purchased products, etc.) to drive innovation and scientific discovery. Unfortunately, the number of appropriate datasets is limited because providing such data comes with certain risks. Typically, banks are wary of sharing their data due to potential leaks of confidential information or violations of data protection regulations. In addition, the data is considered to have commercial value, and companies do not want to disclose it. To mitigate these risks, all necessary steps must be taken to ensure data security and remove any identifying information. This allows information to be shared without violating client confidentiality, but this procedure requires significant effort from engineers, managers, and lawyers. Though there exist a small number of properly anonymized banking datasets, such as credit card transactions~\cite{padhi2021tabular} or AlphaBattle~\cite{alphabattle}, to the best of our knowledge, there are no publicly available large multimodal temporal datasets for banks. 

Thus, this paper introduced the first large-scale multimodal banking dataset to support future research on multimodal techniques for event sequences. In particular, we select several practically essential tasks, such as campaigning~\cite{10.1145/502512.502581}, i.e., prediction if a client would purchase some of four rather popular products in the next month. Each client is described by sequences of typical bank data: transactions, geo positions where the customer used the bank application, and dialogues with technical support. These data sources highlight the main difficulties in developing multimodal models: asynchronous events in different modalities, various intensities of events, rare/irregular events, and the absence of some modalities for many clients. Based on our dataset, the researchers can study cross-modal connections of sequences from multiple sources at the level of individual events.

Let us discuss the details of the dataset collection procedure. At first, we select a complete sample of clients for two years (2021 and 2022) to cover all seasons. Among all customers who had the opportunity to purchase at least one of four products during 2022, we randomly chose 2,186,230 clients, among which 1M customers are labeled by monthly aggregated purchases of each of four products in each month. For these clients, we collect 947,899,612 financial operations, 1,117,213,760 geo position events, and 5,080,781 dialogues with technical support. 
Many clients do not have all three modalities simultaneously because they can never make a transaction, call tech support, or leave their geo trace while running the bank application. 
Next, all the data are properly anonymized to guarantee the confidentiality and privacy of customer information. As a result, it is impossible to recover real clients from our anonymous data. However, we will experimentally show that such anonymization still allows us to extract valuable client information.

We present a comparative analysis of various event sequence datasets alongside the MBD dataset in Table~\ref{table:datasets_stat}. Our dataset is substantially more comprehensive, offering more modalities, events, clients, and downstream tasks than other datasets. MBD incorporates diverse modalities, including bank transactions, geo-locations, and technical support dialogues, providing a richer and more realistic basis for analysis. Additionally, it supports a broader range of downstream tasks, which are detailed in the following sections, allowing for more sophisticated and flexible modeling approaches.

Given the substantial computational resources required for research with our dataset, we have curated a smaller and more balanced subset of MBD (``MBD-small'') to facilitate experimentation under resource-constrained conditions. We included all clients with at least one positive target value from the last month. Then, we also sampled clients with all zeros in their targets at a ratio of 1:10. The balanced subset contains 69474 clients in total. This subset retains the key characteristics of the full dataset while significantly reducing computational overhead. 

To demonstrate data transformation from various sources, Fig.~\ref{fig:schema-downstream} briefly shows examples for each modality, and the temporal data structure for a campaigning model is presented. Let us briefly describe each modality in the following subsection. Further details are provided in Appendix~\ref{sec:appendix_stat}, including the sequence length of event sequences and data samples. 

\subsection{Modalities}

\textbf{1. Bank transactional data} are financial operations carried out between different clients. The main component of our MBD dataset is the client's transactional history, represented by an event with a timestamp and various attributes of the anonymized counterparty. Clients have 638 transactions on average. Collected over two years (2021 and 2022), the sequence of financial operations can uniquely characterize the client~\cite{babaev2022coles}, so this data source plays one of the most significant roles in planning and recommendations.

\textbf{2. Dialog data} consists of transcriptions from customer calls to technical support and negotiations between clients and their managers. We incorporate dialogues from key communication channels, including sales and service calls, which account for most interactions with bank customers. It is an extremely important source of information about client needs and problems~\cite{bauman2022long}. The audio utterance is fed into a commercial Speech-to-Text library. Personal information, e.g., the client's name, is detected in the text and masked. To further anonymize the dialogue, we feed its text into a pre-trained NLP model\footnote{ https://huggingface.co/ai-forever/ruBert-base} and save the resulting embeddings of size 768 in dialogue modality. Only 46\% of customers contact support and have records of conversations, and 98\% of them have no more than 10 dialogues.

\textbf{3. Geostream data} contains a sequence of geo-coordinates of a client obtained throughout 2022. To anonymize this modality, the coordinates are encoded using geohashes\footnote{https://pypi.org/project/pygeohash/}, a geocoding system that converts a geographic location into a short string. Each unique geohash corresponds to a region on the Earth's surface. It is possible to adjust the accuracy and size by removing characters from the end of the code. In our dataset, the coordinates are encoded with a precision of 4, 5, and 6 characters, representing cells of different sizes on the map. As a result, there are 43,999 distinct values of geohash\_4, 347,698 numbers of geohash\_5, and 2,264,404 most precise locations (geohash\_6).

\textbf{4. Products purchases}.
Our dataset serves as a valuable resource for analyzing bank customers' needs and optimizing the campaigning process. This critical task directly impacts the bank's product volume and overall profitability. High-quality recommendations play a key role in enhancing the customer experience, making campaigning essential for business outcomes and customer satisfaction. 
Specifically, MBD includes monthly data on the purchases of four distinct banking products throughout 2022, providing a broader temporal scope that captures patterns beyond the pandemic's peak. We concentrate on the most popular of these products, as internal analysis across various tasks using proprietary data consistently showed that these products provide a robust foundation for model selection. The insights from this data enable the development of models that demonstrate superior performance across a broad spectrum of related tasks. To predict a purchase in a certain month, it is necessary to take events (transactions, geo, dialogues) strictly before the beginning of this month. Therefore, the purchase dataset's date range is shifted by 1 month, i.e., information is available from February 1, 2022, to January 31, 2023. The campaigning task is a multi-label classification problem, i.e., we store a binary label for each product that indicates whether a customer purchases it in a certain month. 
The peculiarity of this dataset is its imbalance, which is specific to this type of business task: 81\% of clients have no purchases, 15\% have one, and the remaining 4\% have two or more purchases. A historical overview over 12 months allows us to model the customer behavior dynamic and predict the date of purchase more accurately. 

\subsection{Data anonymization}\label{sec:dataset-anonymization}

Our dataset contains no personal or confidential information whatsoever. Nevertheless, the event sequences are detailed enough that it could be possible to compare individuals from the publicly accessible portion of the dataset with the original proprietary data. To mitigate this risk, noise is introduced to the data, ensuring that such comparisons and identification are impossible. Our bank's internal security department selected the noise patterns. These patterns are applied locally, preserving the overall structure of the data. The specific noise parameters are not disclosed to prevent potential attacks on the dataset.

All ID fields are hashed with a random salt. All categorical field values are mapped to enumerated indexes. Random noise is added to numerical fields and dates, preserving the hour of the original date, which may cause the local sequence's shuffle. The dialogue embedding space is divided into regions, which are then shuffled.

In the experimental study for downstream tasks, we demonstrate that the applied anonymization techniques have a minimal impact on the ranking of models trained on our dataset. This indicates that the anonymization process preserves fundamental structural properties while retaining information essential for the downstream task, thereby ensuring the reliability of model evaluation.

\section{Benchmark}\label{sec:benchmark}
In this paper, we introduce a benchmark for widely used event sequence datasets, incorporating practically important downstream tasks. This section provides a detailed description of the downstream tasks, baseline methods, and evaluation protocols.

\subsection{Datasets and downstream tasks}\label{sec:benchmark-datasets}

We implement an out-of-fold validation protocol for each downstream task in every dataset. The client dataset is partitioned into five folds, with four folds used for training and the remaining fold reserved for testing. The training and testing sets are publicly available alongside the dataset, allowing future researchers to compare performance metrics. As each dataset in our benchmark exhibits label imbalance, ROC AUC is the most robust and informative evaluation metric due to its resilience to class imbalances. In real-world business, campaign effectiveness is measured by revenue, but conducting A/B tests for every ML model is impractical. Instead, ROC AUC is a reliable proxy metric for model comparison, with only the top performers advancing to A/B testing. This method was validated through real-world A/B tests and is recognized as the core evaluation metric by leading institutions, including one of the largest global banks. The choice of AUC for campaigning is supported by the reason that ranking models by their ROC curves are similar to comparing their non-response ratio at all possible cutoff points simultaneously~\cite{10.1145/2124295.2124353,10.1145/502512.502581}. Let us discuss the details of downstream tasks for each dataset in our benchmark.

\textbf{1. MBD.} For our dataset, we introduce a campaigning downstream task~\cite{essay101228, balyemah2024predicting, hendriksen2020analyzingpredictingpurchaseintent}. In this task, it is required to predict the customer's propensity to purchase four different products in the next month (Fig.~\ref{fig:schema-downstream}) given sequences of transactions, geo locations, and dialogues from the beginning of this month. Solutions to this problem are used to plan marketing campaigns and prepare sales communications through various channels with the client. 

The baseline methods outlined in the following section are applied. The multi-label classification models are trained using our training set. Considering the temporal structure of our target, we compute the embedding of each client's history for up to one month, focusing on the presence of the target product (Fig.~\ref{fig:schema-downstream}). We then evaluate the model using the mean ROC AUC across four binary product labels over the 12 months of 2022.

\textbf{2. Datafusion.}
In this dataset, the downstream task is to predict the higher education attainment of bank clients. It involves analyzing two modalities (transaction histories and clickstream data) to accurately infer client's educational background. The task is formulated as a binary classification problem, with 75\% of the labels corresponding to clients with higher education. 

\textbf{3. Alphabattle.} 
We incorporate the large unimodal Alphabattle dataset alongside the multimodal datasets into our benchmark. This inclusion of data from various financial institutions aims to support more robust and reliable conclusions. In this dataset, the downstream task estimates the probability of a customer defaulting based on their historical card transaction behavior. This task is framed as a binary classification problem, with 2.7\% of the labels representing clients who have defaulted. Although the dataset is unimodal, the downstream task remains highly relevant for financial institutions, offering critical insights into credit risk assessment in improving decision-making processes related to customer management and financial strategies.

\textbf{Multimodal matching.} For the multimodal datasets, MBD and Datafusion, we propose a downstream task of multimodal matching~\cite{zong2023selfsupervised, 10.1145/3459637.3482308, taghibakhshi2023hierarchical, 10.1145/3077136.3080682}. It involves aligning and comparing modalities to identify meaningful relationships or connections. Data from multiple sources for the same client are frequently matched using predefined rules or heuristics, which may not always yield optimal results. To improve its accuracy, specialized identification algorithms are required to compare modalities more precisely.

For the matching task, we employ a framework analogous to CLIP~\cite{radford2021learningtransferablevisualmodels}. We utilize GRU encoders to embed pairs of samples from two input modalities, labeling them as either positive (i.e., data from the same client) or negative matches (i.e., data from different clients). The model is trained using the InfoNCE loss function~\cite{chen2020simpleframeworkcontrastivelearning}, which maximizes similarity for positive pairs while minimizing it for negative pairs. We use Recall@1, Recall@50, and Recall@100 metrics to assess the model's performance.

\subsection{Methods}

To establish performance baselines, we implement several widely adopted architectures.  Our approach prioritizes unsupervised and semi-supervised methods~\cite{balestriero2023cookbook}, enabling the training of a general-purpose encoder on unlabeled sequential data. Additionally, we incorporate supervised methods that allow for the immediate training of the encoder in a fully supervised manner.

\subsubsection{Unimodal approaches}

The following techniques are implemented in our benchmark to extract features from data. First, \textbf{Aggregation Baseline} that contains hand-crafted aggregation statistics~\cite{babaev2022coles}. 
Events are represented either numerically, such as transaction amount, or categorically, like event types. For numerical attributes, we apply aggregation functions (e.g., sum, mean, std) across all events in a sequence. Categorical attributes are grouped by unique values, aggregating numerical attributes using functions like count or mean.

Second, \textbf{CoLES} (Contrastive Learning for Event Sequences), a self-supervised contrastive model~\cite{babaev2022coles} specially developed to obtain representations of such event sequences as bank transactions. The sequence encoder is a GRU (Gated Recurrent Unit) with a hidden size 256. Despite its simplicity, CoLES continues to demonstrate top-level performance in event sequence analysis~\cite{osin2024ebeseasybenchmarkingevent}.

Finally, two \textbf{Tabular Transformers} from IBM~\cite{padhi2021tabular}: 1) TabBERT that adapts BERT to event sequences such as bank transactions, and 2) TabGPT that was initially proposed to generate synthetic tabular sequences. Both models extract 256-dimensional embeddings of an input event sequence. After that, we pool output embeddings of the client in result embedding of size 1024, calculating min, max, mean, and std.

To obtain a representation of a sequence of \textbf{dialogues}, we borrow conventional techniques to aggregate the sequence of embeddings of each client's dialog: 1) mean pooling of all embeddings, and 2) use only the most recent embedding for the date of interest.

For unimodal supervised methods (Supervised RNN), we utilize GRU architectures~\cite{babaev2019rnn} with a hidden size of 32. The models are trained in a multi-label setting using binary cross-entropy (BCE) loss, ensuring effective optimization for tasks with multiple targets.

\subsubsection{Multimodal approaches}

To explore the potential of \textbf{multimodal processing} for event sequence analysis, we compare several fusion techniques: Blending, Late, and Early Fusion. \textbf{Blending} computes a weighted sum of class posterior probabilities from individual single-modal classifiers, effectively combining the predictions from each modality.

In \textbf{Late Fusion}, embeddings from all data sources are concatenated and fed into a classifier. This technique allows the model to learn interactions between modalities after they have been individually processed~\cite{fuse2020med}. In supervised Late Fusion, we utilize separate GRU encoders for each modality, concatenating their embeddings to form a unified representation (Supervised RNN).
    
\textbf{Early Fusion} combines representations from multiple modalities at the initial stages of the model, enabling joint processing of multimodal data. We employ the CrossTransformer approach~\cite{zhang2023crossformer}, which utilizes a cross-attention mechanism to integrate information across modalities efficiently. In our experiments, this method is applied within a supervised learning framework.

\subsubsection{LLM approaches}

In addition to the methods described above, we also explore the use of Large Language Models (LLMs) for multimodal data analysis. Leveraging their ability to process sequential data through universal text-based representations, LLMs offer a versatile approach to analyzing events across different modalities. 

We used the X-separated serialization ~\cite{fang2024large} to represent transactions in text format. X-separated serialization formats the data such that each transaction is represented as a separate line (row), with individual features separated by tabs (columns). In an unimodal setup, we limited transactions to 256 events and four features (event\_time, event\_type, amount, src\_type3). In the multimodal setup, we used 128 transactions and 64 geo events. Serialized transactions and geo hashes are placed one after the other and separated. We used a special delimiter to separate modalities. 


We applied LLMs on serialized data, considering each user as a distinct document. Transactional sequence embedding is obtained from the last hidden layer's last token. No extra model pretraining or prompting has been applied. These embeddings were used to solve downstream tasks for each target. We also used PCA on the embeddings to reduce their dimensionality to 256, matching the baseline embeddings. Our findings indicate that applying PCA to text embeddings reduces dimensionality and improves model predictive performance on downstream tasks. We used a popular AutoML library, LightAutoML ~\cite{vakhrushev2021lightautoml} to solve downstream tasks. It provides a fully automatic pipeline for model selection and hyperparameter tuning.

\begin{table*}
\caption{Mean ROC-AUC of downstream results using unimodal methods.}
\label{table:unimodal methods}
\begin{tabular}{lcccccc}
\toprule
{\textbf{Model}} & \multicolumn{2}{c}{\textbf{MBD}} & \multicolumn{2}{c}
{\textbf{Datafusion}} & \multicolumn{1}{c}{\textbf{Alphabattle}}\\ 
 & \textbf{Transactions} & \textbf{Geostream} & \textbf{Transactions} & \textbf{Clickstream} & \textbf{Transactions} \\ 
\midrule
\textbf{Aggregation} & 0.783 $\pm$ 0.002 & 0.595 $\pm$ 0.002 & 0.793 $\pm$ 0.013 & 0.537 $\pm$ 0.018& 0.785 $\pm$ 0.0010\\ 
\textbf{CoLES} & 0.773 $\pm$ 0.002 & 0.598 $\pm$ 0.004 & 0.784 $\pm$ 0.012& 0.641 $\pm$ 0.013& 0.793 $\pm$ 0.0005\\ 
\textbf{TabBERT} & 0.762 $\pm$ 0.004 & 0.603 $\pm$ 0.002 & 0.762 $\pm$ 0.014 & 0.590 $\pm$ 0.026 & 0.778 $\pm$ 0.0003\\ 
\textbf{TabGPT} & 0.802 $\pm$ 0.002& 0.621 $\pm$ 0.003 & 0.766 $\pm$ 0.013 & 0.618 $\pm$ 0.016& 0.775 $\pm$ 
 0.0010 \\
\textbf{Supervised RNN} & 0.819 $\pm$ 0.002 & 0.540 $\pm$ 0.012 & 0.712 $\pm$ 0.016 & 0.563 $\pm$ 0.011 & 0.792 $\pm$ 0.0030 \\ 
\bottomrule
\end{tabular}
\end{table*}

\begin{table*}
 \caption{Mean ROC-AUC in late fusion setting.}
\label{table:multimodal methods}
 \begin{tabular}{llcccc}
\toprule
 \textbf{Dataset} & \textbf{Modalities} & \textbf{CoLES} & \textbf{TabGPT} & \textbf{TabBERT} & \textbf{Supervised RNN}\\ \midrule
 \textbf{MBD} & Trx & 0.773 $\pm$ 0.002 & 0.802 $\pm$ 0.001 & 0.762 $\pm$ 0.004 & 0.819 $\pm$ 0.002 \\
& Trx + Geo &0.775 $\pm$ 0.002 & 0.800 $\pm$ 0.001 & 0.764 $\pm$ 0.004 & 0.819 $\pm$ 0.001\\
& Trx + Dialog &0.781 $\pm$ 0.002 & 0.810 $\pm$ 0.002 & 0.773 $\pm$ 0.003 & 0.821 $\pm$ 0.0006 \\
& Trx + Dialog + Geo &0.783 $\pm$ 0.002 & 0.808 $\pm$ 0.001 & 0.775 $\pm$ 0.003 & 0.824 $\pm$ 0.001\\
\hline
\textbf{Private dataset} & Trx &0.772 $\pm$ 0.002 & 0.796 $\pm$ 0.000 & 0.754 $\pm$ 0.011 & 0.813 $\pm$ 0.002 \\
& Trx + Geo & 0.772 $\pm$ 0.002 & 0.796 $\pm$ 0.001 & 0.756 $\pm$ 0.011 & 0.815 $\pm$ 0.001 \\
& Trx + Dialog & 0.776 $\pm$ 0.002 & 0.805 $\pm$ 0.001 & 0.765 $\pm$ 0.009 & 0.817 $\pm$ 0.0006 \\
& Trx + Dialog + Geo & 0.777 $\pm$ 0.002 & 0.802 $\pm$ 0.001 & 0.766 $\pm$ 0.009 & 0.821 $\pm$ 0.001 \\
\hline
\textbf{Datafusion} & Trx & 0.784 $\pm$ 0.012 & 0.766 $\pm$ 0.013& 0.762 $\pm$ 0.014 & 0.712 $\pm$ 0.015 \\
& Trx + Click &0.785 $\pm$ 0.011 & 0.766 $\pm$ 0.011 & 0.761 $\pm$ 0.012 & 0.703 $\pm$ 0.008 \\
\bottomrule
\end{tabular}
\end{table*}

\section{Experiments}\label{sec:exper}

Our models, experiments, and training procedures were implemented in Python, leveraging PyTorch and PyTorch Lightning for deep learning tasks, and PySpark for distributed data processing. We trained the models on  NVIDIA V100 GPU, while the gradient boosting models were trained on computational clusters equipped with 600 CPU cores. 

\subsection{Model and training hyperparameters}\label{sec:model-training-hyperparameters}

We employ the unsupervised baseline methods (CoLES, TabGPT, TabBERT, Aggregation) with default hyperparameters from the pytorch-lifestream framework. 
The Adam optimizer is utilized, with an initial learning rate of 0.001, coupled with the StepLR scheduler. Models are saved based on the lowest validation loss or the highest validation unsupervised metrics \cite{tsitsulin2023unsupervised}, evaluated after each training epoch, with training of 15 epochs. We employ 24-dimensional embeddings for categorical features and clipped the number of categories for features with many unique values. We apply either an identical mapping or a logarithmic transformation for numerical features. We use the gradient boosting algorithm from PySpark ML for downstream tasks.

\subsection{Downstream tasks}

\subsubsection{MBD: campaigning task} This Subsection contains experimental results for the campaigning task in the MBD dataset in unimodal and multimodal baselines. One of the main objectives of our paper is to provide a real data benchmark to facilitate the development of multimodal algorithms. It is necessary to demonstrate that an algorithm outperforming another on our public benchmark will similarly outperform it on real data. 

Table \ref{table:unimodal methods} shows that the transaction modality is crucial in achieving accurate classification. In contrast, dialogues and geostream perform only slightly better than a random estimator when used in unimodal setting. However, as shown in Table \ref{table:multimodal methods}, the predictive performance improved significantly in the multimodal setting by integrating additional modalities. The overall trend indicates a consistent improvement in validation metrics as more modalities are incorporated. Specifically, the multimodal late fusion approach enhances predictive accuracy by 1-1,5\% when adding other data sources to the transaction stream.

\begin{figure}[h]
  \centering
  \includegraphics[width=0.8\columnwidth]{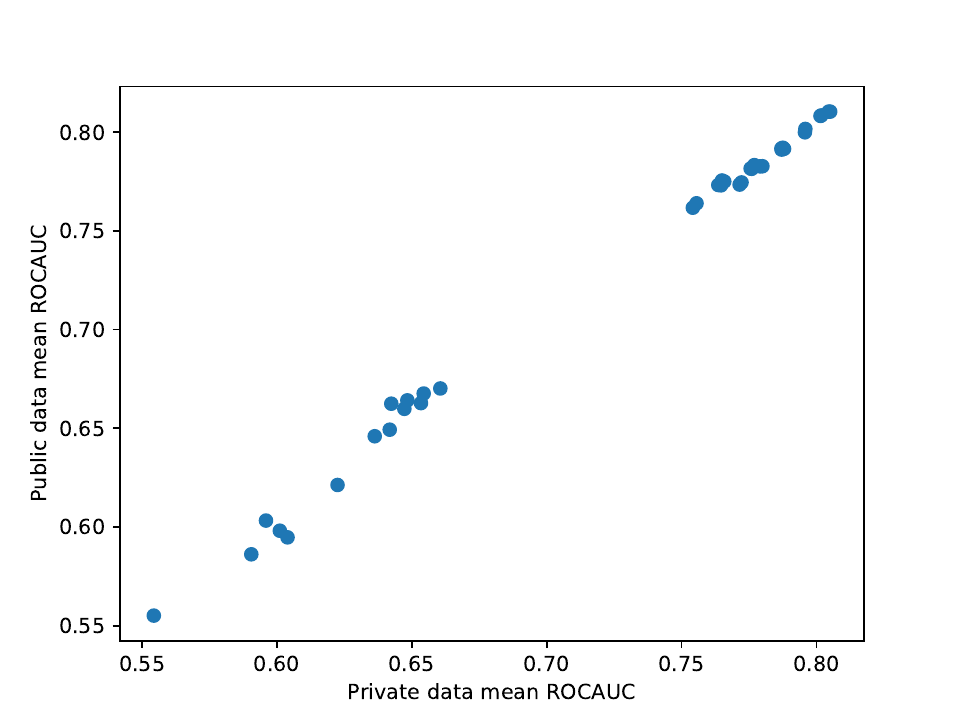}
  \caption{Model performance comparison on private and public data for the campaigning task.Kendall-Tau=0.94}
  \Description{Model performance comparison on private and public data.}
 \label{fig:pp_corr}
\end{figure}

Fig.~\ref{fig:pp_corr} highlights a strong correlation between performance metrics on public and private datasets, with a Kendall-tau correlation coefficient of 0.94. This figure presents performance results derived from the extended experiments detailed in Appendix~\ref{sec:appendix_expers} (Tables~\ref{table:private_fusion}, \ref{table:public_fusion}) for the campaigning task on both datasets. The private dataset is structurally equivalent to the public dataset, maintaining identical client composition, temporal coverage, and data volume, with the sole distinction being the application of anonymization. Hence, the anonymization process has minimal impact on model performance for the downstream task of campaigning. The consistency in relative ranking across both datasets (Table~\ref{table:multimodal methods}) underscores the reliability of our benchmark for advancing research in multimodal event sequence analysis.

More detailed comparison of results on MBD and the private dataset and comprehensive results for each modality and all possible fusion combinations of multiple modalities are shown in Appendix~\ref{sec:appendix_expers} (Tables~\ref{table:private_blending} -~\ref{table:public_fusion}). Our modalities, namely, geostream, transactions, and dialogues, are denoted as Geo, Trx, and Dialogs. We use a clear notation to specify a method applied to a modality. For instance, if embeddings from the CoLES model are applied to transactions, we denote this as TrxCoLES.

\subsubsection{Datafusion: Higher Education}
In this subsection, we present the results of unimodal and multimodal experiments on the DataFusion dataset, as shown in Tables~\ref{table:unimodal methods} and \ref{table:multimodal methods}. While the dataset is small, leading to an insignificant increase in quality metrics when incorporating additional sources in multimodal settings, the results in both unimodal and multimodal configurations remain valuable as they contribute to expanding our benchmark.

\subsubsection{Alphabattle: Default}

In Table \ref{table:unimodal methods}, we present the metrics for unimodal methods on the Alphabattle dataset. Interestingly, the supervised methods on both the MBD and Alphabattle datasets perform as well as, or better than, unsupervised methods. This may be attributed to the dataset size and the available labeled data.

\subsubsection{Multimodal matching task}

In this subsection, we present the results of our proposed multimodal matching benchmark, summarized in Table~\ref{table:matching_part}, which includes both MBD and Datafusion datasets. Detailed results for other modalities can be found in Appendix~\ref{sec:appendix_expers}, in Table~\ref{table:matching}. 
We report Recall@1, Recall@50, and Recall@100 for each modality, measured in both directions.  We also present the multimodal matching results for the DataFusion dataset. For example, in the case of the transaction and geostream pair for MBD, we compute both Trx2Geo and Geo2Trx, evaluating alignment from both perspectives. 
Our analysis reveals considerable variation in performance across different modality pairs. Specifically, dialogues consistently exhibit weaker matching performance than other modalities, such as transactions and geostream, demonstrating significantly stronger alignment. This disparity suggests potential limitations within the dialogue modality, indicating that it may offer less complementary or aligned information. Alternatively, the unique structure of dialogue data may necessitate more sophisticated techniques for effective integration with other modalities. 

\begin{table}
\caption{Multimodal matching task}
 \label{table:matching_part}
 \centering
  \resizebox{\columnwidth}{!}{
 \begin{tabular}{llccc}
 \toprule
\textbf{Dataset}&\textbf{Modalities}&\textbf{Recall@1}&\textbf{Recall@50}&\textbf{Recall@100} \\
\midrule
\textbf{MBD} & Trx2Geo&0.006 $\pm$ 0.0003&0.196 $\pm$ 0.002& 0.303 $\pm$ 0.004 \\
&Geo2Trx&0.004 $\pm$ 0.0003&0.162 $\pm$ 0.002&0.262 $\pm$ 0.004 \\
\hline
\textbf{Datafusion} &Trx2Click&0.002 $\pm$ 0.0010&0.063 $\pm$ 0.004&0.120 $\pm$ 0.009 \\
&Click2Trx&0.001 $\pm$ 0.0007&0.070 $\pm$ 0.005&0.115 $\pm$ 0.008 \\
\bottomrule

\end{tabular}
}
\end{table} 


\begin{table*}
    \caption{Comparison of Multimodal Fusion Techniques: Blending, Late Fusion, and Early Fusion.}
    \label{table:multimodal_adv_methods}
  \begin{tabular}{ll|c|cc|c}
    \toprule
    \textbf{Dataset} & \textbf{Modalities} & \multicolumn{1}{c|}{\textbf{Blending}} & \multicolumn{2}{c|}{\textbf{Late Fusion}} & \textbf{Early Fusion} \\ 
            & & \textbf{TabGPT} & \textbf{TabGPT} & \textbf{Supervised RNN} & \textbf{CrossTransformer} \\
            \midrule
    \textbf{MBD} & Trx + Geo & 0.804 $\pm$ 0.001 & 0.800 $\pm$ 0.001 & 0.819 $\pm$ 0.001 & 0.815 $\pm$ 0.001 \\ 
            & Trx + Dialog & 0.742 $\pm$ 0.001 & 0.810 $\pm$ 0.002 & 0.821 $\pm$ 0.0006 & 0.821 $\pm$ 0.002 \\ 
            \hline
            \textbf{Datafusion} & Trx + Click & 0.756 $\pm$ 0.013 & 0.766 $\pm$ 0.011 & 0.703 $\pm$ 0.008 & 0.735 $\pm$ 0.010 \\ 
  \bottomrule
\end{tabular}
\end{table*}

\subsection{Comparison of Multimodal Fusion Methods}

In this subsection, we evaluate the performance of multimodal fusion techniques across multiple datasets. For this analysis, we select the best-performing models for Blending and Late Fusion to compare against Early Fusion. As shown in Table~\ref{table:multimodal_adv_methods}, Late Fusion with a Supervised RNN consistently delivers superior results on the MBD dataset, demonstrating its robustness in effectively integrating multimodal data. Early Fusion, implemented via the cross-attention from CrossTransformer~\cite{zhang2023crossformer}, achieves competitive performance but slightly lags behind Late Fusion, while Blending exhibits considerably weaker results. On the Datafusion dataset, Late Fusion with TabGPT is the most effective method, with Early Fusion also performing well and outperforming Late Fusion with a Supervised RNN. Blending, in contrast, consistently yields the lowest performance across all modalities. These findings highlight the effectiveness and reliability of Late Fusion for multimodal data integration while identifying Early Fusion as a promising direction for further research and optimization.

\subsection{LLMs for multimodal banking data}\label{sec:llm}

Large Language Models, or LLMs, are now strong tools for handling various types of tabular data ~\cite{ruan2024language}. This includes sequential tabular data ~\cite{zhang2023fata}. We used popular models LLAMA3.2 ~\cite{meta2024llama} and QWEN2.5 ~\cite{yang2024qwen2}. These two are publicly available. We also used two in-house LLMs, Gigar4 and Gigar4 SFT (Supervised Fine-Tuning). The last two models are Russian language-specific language models. Gigar4 SFT is a version that has been fine-tuned on several math datasets. The fine-tuning of math datasets enhances their ability to process and predict numerical data accurately. We took 3B parameter versions of the models to keep inference time low while being able to process context up to 8192 tokens on a single GPU.

Experiments with LLMs are computationally expensive, so we used the MBD-small subset described in Subsection~\ref{subsec:mbd_overview}. Inference on this dataset using 3B parameters LLMs takes 12 hours on a single NVIDIA A100 GPU. Our results (Table~\ref{tab:llm}) show that in transactions, relatively small language models can match the performance of specialized event-sequence models, even without extra tuning. At the same time, there is room for improvement to better account for data-specific characteristics. Geo hashes exhibit extreme diversity in values (44K unique anonymized values even for level-4 hashes), which makes it challenging for models to generalize without fine-tuning. Additionally, adding modalities increases the required context length, which is already a challenge for LLMs, given the long event sequences and the way they are encoded into text.

\begin{table}
  \caption{Comparison of LLM-based approaches and Multimodal Fusion  for MBD-small dataset}
  \label{tab:llm}
  \resizebox{\columnwidth}{!}{
  \begin{tabular}{ll|c|c|c}
    \toprule
    \textbf{} & \textbf{Model} & \textbf{Trx} & \textbf{Trx+Geo} & \textbf{Geo} \\
    \midrule
    & Gigar4  & 0.773$\pm$0.007 & 0.737$\pm$0.006 & 0.548$\pm$0.004 \\ 
              \textbf{LLM} & Gigar4 SFT & 0.778$\pm$0.005 & 0.735$\pm$0.014 & 0.549$\pm$0.010 \\ 
                & LLAMA3.2 & 0.773$\pm$0.006 & 0.735$\pm$0.011 & 0.534$\pm$0.016 \\ 
                 & QWEN2.5 & 0.757$\pm$0.005 & 0.727$\pm$0.007 & 0.539$\pm$0.008 \\ 
    \hline
& CoLES &  0.785$\pm$0.006 & 0.782$\pm$0.005 & 0.568$\pm$0.007 \\ 
                  & TabGPT &  0.779$\pm$0.007 & 0.782$\pm$0.002 & 0.594$\pm$0.016 \\ 
                 \textbf{Baselines} & TabBERT & 0.781$\pm$0.003 & 0.789$\pm$0.002 & 0.601$\pm$0.013 \\ 
             & Aggregation & 0.781$\pm$0.009 & 0.783$\pm$0.008 & 0.530$\pm$0.006 \\
             & Supervised RNN & 0.745$\pm$0.004 & 0.742$\pm$0.006 & 0.532$\pm$0.013 \\
  \bottomrule
\end{tabular}
}
\end{table}

\section{Conclusion and Future Work}\label{sec:conc}

In this paper, we introduce MBD, the first large-scale, publicly available multimodal banking dataset (Table~\ref{table:datasets_stat}), comprising sequential data from over 2 million clients, including transactions, geo-locations, and support dialogues. MBD ensures compliance with data protection regulations, enabling research on financial data while maintaining privacy. Our findings show that anonymization has minimal impact on algorithm performance (Table~\ref{table:multimodal methods}, Fig.~\ref{fig:pp_corr}), making the dataset suitable for real-world deployment. Additionally, excluding sensitive attributes (e.g., gender, age, race) mitigates bias, fostering ethical AI development.

MBD, along with Datafusion and Alphabattle, establishes a benchmark for key downstream tasks, facilitating the development of scalable multimodal and unimodal algorithms for event sequence prediction. We implemented the state-of-the-art models in banking domains, demonstrating that even basic multimodal fusion outperforms single-modal baselines (Table~\ref{table:multimodal methods}). Further, Late and Early Fusion experiments (Table~\ref{table:multimodal_adv_methods}) highlight the potential for improving model performance by effectively capturing cross-modal interactions. We also assess LLMs in unimodal and multimodal settings, comparing them with domain-specific approaches (Table~\ref{tab:llm}). Given MBD’s scale and practical relevance, even modest performance gains can yield significant financial benefits.

Future iterations will incorporate additional data sources (e.g., clickstreams, enriched dialogue features) and expand downstream tasks to financial product recommendations, customer churn prediction, fraud detection, and credit risk assessment. Extending the dataset’s time range will further enable longitudinal studies and trend analysis


\bibliographystyle{ACM-Reference-Format}
\bibliography{refs}


\begin{thebibliography}{60}


\ifx \showCODEN    \undefined \def \showCODEN     #1{\unskip}     \fi
\ifx \showISBNx    \undefined \def \showISBNx     #1{\unskip}     \fi
\ifx \showISBNxiii \undefined \def \showISBNxiii  #1{\unskip}     \fi
\ifx \showISSN     \undefined \def \showISSN      #1{\unskip}     \fi
\ifx \showLCCN     \undefined \def \showLCCN      #1{\unskip}     \fi
\ifx \shownote     \undefined \def \shownote      #1{#1}          \fi
\ifx \showarticletitle \undefined \def \showarticletitle #1{#1}   \fi
\ifx \showURL      \undefined \def \showURL       {\relax}        \fi
\providecommand\bibfield[2]{#2}
\providecommand\bibinfo[2]{#2}
\providecommand\natexlab[1]{#1}
\providecommand\showeprint[2][]{arXiv:#2}

\bibitem[age({[n.\,d.]})]%
        {agepred}
 \bibinfo{year}{[n.\,d.]}\natexlab{}.
\newblock \bibinfo{title}{Age Prediction Dataset}.
\newblock \bibinfo{howpublished}{\url{https://ods.ai/competitions/sberbank-sirius-lesson}}.
\newblock
\newblock
\shownote{Accessed: 2024-06-07}.


\bibitem[alp({[n.\,d.]})]%
        {alphabattle}
 \bibinfo{year}{[n.\,d.]}\natexlab{}.
\newblock \bibinfo{title}{Alfa {Battle} 2.0 competition}.
\newblock \bibinfo{howpublished}{\url{https://boosters.pro/championship/alfabattle2/overview}}.
\newblock
\newblock
\shownote{Accessed: 2024-06-07}.


\bibitem[ame({[n.\,d.]})]%
        {amex}
 \bibinfo{year}{[n.\,d.]}\natexlab{}.
\newblock \bibinfo{title}{American Express - Default Prediction}.
\newblock \bibinfo{howpublished}{\url{https://www.kaggle.com/competitions/amex-default-prediction}}.
\newblock
\newblock
\shownote{Accessed: 2024-06-07}.


\bibitem[dat({[n.\,d.]})]%
        {datafusion2022}
 \bibinfo{year}{[n.\,d.]}\natexlab{}.
\newblock \bibinfo{title}{Data Fusion Contest 2022}.
\newblock \bibinfo{howpublished}{\url{https://ods.ai/tracks/data-fusion-2022-competitions}}.
\newblock
\newblock
\shownote{Accessed: 2024-06-07}.


\bibitem[ros({[n.\,d.]})]%
        {rosbank}
 \bibinfo{year}{[n.\,d.]}\natexlab{}.
\newblock \bibinfo{title}{Rosbank ML competition}.
\newblock \bibinfo{howpublished}{\url{https://boosters.pro/championship/rosbank1/overview}}.
\newblock
\newblock
\shownote{Accessed: 2024-06-07}.


\bibitem[ama(2018)]%
        {amazon}
 \bibinfo{year}{2018}\natexlab{}.
\newblock \bibinfo{title}{{Amazon Product Reviews}}.
\newblock \bibinfo{howpublished}{\url{https://cseweb.ucsd.edu/~jmcauley/datasets.html\#amazon_reviews}}.
\newblock


\bibitem[Ala’raj et~al\mbox{.}(2022)]%
        {ala2022deep}
\bibfield{author}{\bibinfo{person}{Maher Ala’raj}, \bibinfo{person}{Maysam~F Abbod}, \bibinfo{person}{Munir Majdalawieh}, {and} \bibinfo{person}{Luay Jum’a}.} \bibinfo{year}{2022}\natexlab{}.
\newblock \showarticletitle{A deep learning model for behavioural credit scoring in banks}.
\newblock \bibinfo{journal}{\emph{Neural Computing and Applications}} \bibinfo{volume}{34}, \bibinfo{number}{8} (\bibinfo{date}{April} \bibinfo{year}{2022}), \bibinfo{pages}{5839--5866}.
\newblock
\href{https://doi.org/10.1007/s00521-021-06695-z}{doi:\nolinkurl{10.1007/s00521-021-06695-z}}


\bibitem[Babaev et~al\mbox{.}(2022)]%
        {babaev2022coles}
\bibfield{author}{\bibinfo{person}{Dmitrii Babaev}, \bibinfo{person}{Nikita Ovsov}, \bibinfo{person}{Ivan Kireev}, \bibinfo{person}{Maria Ivanova}, \bibinfo{person}{Gleb Gusev}, \bibinfo{person}{Ivan Nazarov}, {and} \bibinfo{person}{Alexander Tuzhilin}.} \bibinfo{year}{2022}\natexlab{}.
\newblock \showarticletitle{Coles: Contrastive learning for event sequences with self-supervision}. In \bibinfo{booktitle}{\emph{Proceedings of the International Conference on Management of Data (SIGMOD)}}. \bibinfo{pages}{1190--1199}.
\newblock


\bibitem[Babaev et~al\mbox{.}(2019)]%
        {babaev2019rnn}
\bibfield{author}{\bibinfo{person}{Dmitrii Babaev}, \bibinfo{person}{Maxim Savchenko}, \bibinfo{person}{Alexander Tuzhilin}, {and} \bibinfo{person}{Dmitrii Umerenkov}.} \bibinfo{year}{2019}\natexlab{}.
\newblock \showarticletitle{{ET-RNN}: Applying deep learning to credit loan applications}. In \bibinfo{booktitle}{\emph{Proceedings of the 25th {ACM} {SIGKDD} international conference on knowledge discovery \& data mining}} (Anchorage AK USA). \bibinfo{publisher}{{ACM}}, \bibinfo{address}{New York, NY, USA}, \bibinfo{pages}{2183--2190}.
\newblock
\href{https://doi.org/10.1145/3292500.3330693}{doi:\nolinkurl{10.1145/3292500.3330693}}


\bibitem[Balestriero et~al\mbox{.}(2023)]%
        {balestriero2023cookbook}
\bibfield{author}{\bibinfo{person}{Randall Balestriero}, \bibinfo{person}{Mark Ibrahim}, \bibinfo{person}{Vlad Sobal}, \bibinfo{person}{Ari Morcos}, \bibinfo{person}{Shashank Shekhar}, \bibinfo{person}{Tom Goldstein}, \bibinfo{person}{Florian Bordes}, \bibinfo{person}{Adrien Bardes}, \bibinfo{person}{Gregoire Mialon}, \bibinfo{person}{Yuandong Tian}, \bibinfo{person}{Avi Schwarzschild}, \bibinfo{person}{Andrew~Gordon Wilson}, \bibinfo{person}{Jonas Geiping}, \bibinfo{person}{Quentin Garrido}, \bibinfo{person}{Pierre Fernandez}, \bibinfo{person}{Amir Bar}, \bibinfo{person}{Hamed Pirsiavash}, \bibinfo{person}{Yann LeCun}, {and} \bibinfo{person}{Micah Goldblum}.} \bibinfo{year}{2023}\natexlab{}.
\newblock \bibinfo{title}{A Cookbook of Self-Supervised Learning}.
\newblock
\showeprint[arxiv]{2304.12210}~[cs.LG]


\bibitem[Balyemah et~al\mbox{.}(2024)]%
        {balyemah2024predicting}
\bibfield{author}{\bibinfo{person}{Abraham~Jallah Balyemah}, \bibinfo{person}{Sonkarlay~JY Weamie}, \bibinfo{person}{Jiang Bin}, \bibinfo{person}{Karmue~Vasco Janda}, {and} \bibinfo{person}{Felix~Jwakdak Joshua}.} \bibinfo{year}{2024}\natexlab{}.
\newblock \showarticletitle{Predicting Purchasing Behavior on E-Commerce Platforms: A Regression Model Approach for Understanding User Features that Lead to Purchasing}.
\newblock \bibinfo{journal}{\emph{International Journal of Communications, Network and System Sciences}} \bibinfo{volume}{17}, \bibinfo{number}{6} (\bibinfo{year}{2024}), \bibinfo{pages}{81--103}.
\newblock


\bibitem[Bauman et~al\mbox{.}(2024)]%
        {bauman2022long}
\bibfield{author}{\bibinfo{person}{Konstantin Bauman}, \bibinfo{person}{Alexey Vasilev}, {and} \bibinfo{person}{Alexander Tuzhilin}.} \bibinfo{year}{2024}\natexlab{}.
\newblock \showarticletitle{Does the Long Tail of Context Exist and Matter? The Case of Dialogue-based Recommender Systems}. In \bibinfo{booktitle}{\emph{Proceedings of the 32nd ACM Conference on User Modeling, Adaptation and Personalization}}. \bibinfo{pages}{273--278}.
\newblock


\bibitem[Baye et~al\mbox{.}(2024)]%
        {Baye2024CustomerRA}
\bibfield{author}{\bibinfo{person}{Irina Baye}, \bibinfo{person}{Tim Reiz}, {and} \bibinfo{person}{Geza Sapi}.} \bibinfo{year}{2024}\natexlab{}.
\newblock \showarticletitle{Customer Recognition and Mobile Geo-Targeting}.
\newblock \bibinfo{journal}{\emph{Review of Industrial Organization}} (\bibinfo{year}{2024}).
\newblock
\urldef\tempurl%
\url{https://api.semanticscholar.org/CorpusID:169092374}
\showURL{%
\tempurl}


\bibitem[Bazarova et~al\mbox{.}(2025)]%
        {bazarova2024universal}
\bibfield{author}{\bibinfo{person}{Alexandra Bazarova}, \bibinfo{person}{Maria Kovaleva}, \bibinfo{person}{Ilya Kuleshov}, \bibinfo{person}{Evgenia Romanenkova}, \bibinfo{person}{Alexander Stepikin}, \bibinfo{person}{Aleksandr Yugay}, \bibinfo{person}{Dzhambulat Mollaev}, \bibinfo{person}{Ivan Kireev}, \bibinfo{person}{Andrey Savchenko}, {and} \bibinfo{person}{Alexey Zaytsev}.} \bibinfo{year}{2025}\natexlab{}.
\newblock \showarticletitle{Learning transactions representations for information management in banks: Mastering local, global, and external knowledge}.
\newblock \bibinfo{journal}{\emph{International Journal of Information Management Data Insights}} \bibinfo{volume}{5}, \bibinfo{number}{1} (\bibinfo{year}{2025}), \bibinfo{pages}{100323}.
\newblock
\showISSN{2667-0968}
\href{https://doi.org/10.1016/j.jjimei.2025.100323}{doi:\nolinkurl{10.1016/j.jjimei.2025.100323}}


\bibitem[Chen et~al\mbox{.}(2020)]%
        {chen2020simpleframeworkcontrastivelearning}
\bibfield{author}{\bibinfo{person}{Ting Chen}, \bibinfo{person}{Simon Kornblith}, \bibinfo{person}{Mohammad Norouzi}, {and} \bibinfo{person}{Geoffrey Hinton}.} \bibinfo{year}{2020}\natexlab{}.
\newblock \bibinfo{title}{A Simple Framework for Contrastive Learning of Visual Representations}.
\newblock
\showeprint[arxiv]{2002.05709}~[cs.LG]
\urldef\tempurl%
\url{https://arxiv.org/abs/2002.05709}
\showURL{%
\tempurl}


\bibitem[Fang et~al\mbox{.}(2024)]%
        {fang2024large}
\bibfield{author}{\bibinfo{person}{Xi Fang}, \bibinfo{person}{Weijie Xu}, \bibinfo{person}{Fiona~Anting Tan}, \bibinfo{person}{Jiani Zhang}, \bibinfo{person}{Ziqing Hu}, \bibinfo{person}{Yanjun~Jane Qi}, \bibinfo{person}{Scott Nickleach}, \bibinfo{person}{Diego Socolinsky}, \bibinfo{person}{Srinivasan Sengamedu}, \bibinfo{person}{Christos Faloutsos}, {et~al\mbox{.}}} \bibinfo{year}{2024}\natexlab{}.
\newblock \showarticletitle{Large language models (LLMs) on tabular data: Prediction, generation, and understanding-a survey}.
\newblock  (\bibinfo{year}{2024}).
\newblock


\bibitem[Fursov et~al\mbox{.}(2021)]%
        {fursov2021gradient}
\bibfield{author}{\bibinfo{person}{Ivan Fursov}, \bibinfo{person}{Alexey Zaytsev}, \bibinfo{person}{Nikita Kluchnikov}, \bibinfo{person}{Andrey Kravchenko}, {and} \bibinfo{person}{Evgeny Burnaev}.} \bibinfo{year}{2021}\natexlab{}.
\newblock \showarticletitle{Gradient-based adversarial attacks on categorical sequence models via traversing an embedded world}. In \bibinfo{booktitle}{\emph{Analysis of Images, Social Networks and Texts: AIST}}. Springer, \bibinfo{pages}{356--368}.
\newblock


\bibitem[Guo et~al\mbox{.}(2020)]%
        {guo2020surveyvisualanalysisevent}
\bibfield{author}{\bibinfo{person}{Yi Guo}, \bibinfo{person}{Shunan Guo}, \bibinfo{person}{Zhuochen Jin}, \bibinfo{person}{Smiti Kaul}, \bibinfo{person}{David Gotz}, {and} \bibinfo{person}{Nan Cao}.} \bibinfo{year}{2020}\natexlab{}.
\newblock \bibinfo{title}{Survey on Visual Analysis of Event Sequence Data}.
\newblock
\showeprint[arxiv]{2006.14291}~[cs.HC]
\urldef\tempurl%
\url{https://arxiv.org/abs/2006.14291}
\showURL{%
\tempurl}


\bibitem[Hao et~al\mbox{.}(2011)]%
        {Hao2011VisualSA}
\bibfield{author}{\bibinfo{person}{Ming~C. Hao}, \bibinfo{person}{Christian Rohrdantz}, \bibinfo{person}{Halld{\'o}r Janetzko}, \bibinfo{person}{Umeshwar Dayal}, \bibinfo{person}{Daniel~A. Keim}, \bibinfo{person}{Lars-Erik Haug}, {and} \bibinfo{person}{Meichun Hsu}.} \bibinfo{year}{2011}\natexlab{}.
\newblock \showarticletitle{Visual sentiment analysis on twitter data streams}.
\newblock \bibinfo{journal}{\emph{2011 IEEE Conference on Visual Analytics Science and Technology (VAST)}} (\bibinfo{year}{2011}), \bibinfo{pages}{277--278}.
\newblock
\urldef\tempurl%
\url{https://api.semanticscholar.org/CorpusID:13022396}
\showURL{%
\tempurl}


\bibitem[Hassan et~al\mbox{.}(2019)]%
        {Hassan2019AutomaticAO}
\bibfield{author}{\bibinfo{person}{Fadi Hassan}, \bibinfo{person}{David S{\'a}nchez}, \bibinfo{person}{Jordi Soria-Comas}, {and} \bibinfo{person}{Josep Domingo-Ferrer}.} \bibinfo{year}{2019}\natexlab{}.
\newblock \showarticletitle{Automatic Anonymization of Textual Documents: Detecting Sensitive Information via Word Embeddings}.
\newblock \bibinfo{journal}{\emph{2019 18th IEEE International Conference On Trust, Security And Privacy In Computing And Communications/13th IEEE International Conference On Big Data Science And Engineering (TrustCom/BigDataSE)}} (\bibinfo{year}{2019}), \bibinfo{pages}{358--365}.
\newblock
\urldef\tempurl%
\url{https://api.semanticscholar.org/CorpusID:207829529}
\showURL{%
\tempurl}


\bibitem[Hayashi et~al\mbox{.}(2019)]%
        {Hayashi2019PreTrainedTE}
\bibfield{author}{\bibinfo{person}{Tomoki Hayashi}, \bibinfo{person}{Shinji Watanabe}, \bibinfo{person}{Tomoki Toda}, \bibinfo{person}{K. Takeda}, \bibinfo{person}{Shubham Toshniwal}, {and} \bibinfo{person}{Karen Livescu}.} \bibinfo{year}{2019}\natexlab{}.
\newblock \showarticletitle{Pre-Trained Text Embeddings for Enhanced Text-to-Speech Synthesis}. In \bibinfo{booktitle}{\emph{Interspeech}}.
\newblock
\urldef\tempurl%
\url{https://api.semanticscholar.org/CorpusID:202737157}
\showURL{%
\tempurl}


\bibitem[Hendriksen et~al\mbox{.}(2020)]%
        {hendriksen2020analyzingpredictingpurchaseintent}
\bibfield{author}{\bibinfo{person}{Mariya Hendriksen}, \bibinfo{person}{Ernst Kuiper}, \bibinfo{person}{Pim Nauts}, \bibinfo{person}{Sebastian Schelter}, {and} \bibinfo{person}{Maarten de Rijke}.} \bibinfo{year}{2020}\natexlab{}.
\newblock \bibinfo{title}{Analyzing and Predicting Purchase Intent in E-commerce: Anonymous vs. Identified Customers}.
\newblock
\showeprint[arxiv]{2012.08777}~[cs.IR]
\urldef\tempurl%
\url{https://arxiv.org/abs/2012.08777}
\showURL{%
\tempurl}


\bibitem[Huang et~al\mbox{.}(2021)]%
        {10.1145/3459637.3482308}
\bibfield{author}{\bibinfo{person}{Hongren Huang}, \bibinfo{person}{Shu Guo}, \bibinfo{person}{Chen Li}, \bibinfo{person}{Jiawei Sheng}, \bibinfo{person}{Lihong Wang}, \bibinfo{person}{Jianxin Li}, \bibinfo{person}{Jing Liu}, {and} \bibinfo{person}{Shenghai Zhong}.} \bibinfo{year}{2021}\natexlab{}.
\newblock \showarticletitle{Two-tier Graph Contextual Embedding for Cross-device User Matching}. In \bibinfo{booktitle}{\emph{Proceedings of the 30th ACM International Conference on Information \& Knowledge Management}} (Virtual Event, Queensland, Australia) \emph{(\bibinfo{series}{CIKM '21})}. \bibinfo{publisher}{Association for Computing Machinery}, \bibinfo{address}{New York, NY, USA}, \bibinfo{pages}{730–739}.
\newblock
\showISBNx{9781450384469}
\href{https://doi.org/10.1145/3459637.3482308}{doi:\nolinkurl{10.1145/3459637.3482308}}


\bibitem[Huang et~al\mbox{.}(2020)]%
        {fuse2020med}
\bibfield{author}{\bibinfo{person}{Shih-Cheng Huang}, \bibinfo{person}{Anuj Pareek}, \bibinfo{person}{Saeed Seyyedi}, \bibinfo{person}{Imon Banerjee}, {and} \bibinfo{person}{Matthew~P Lungren}.} \bibinfo{year}{2020}\natexlab{}.
\newblock \showarticletitle{Fusion of medical imaging and electronic health records using deep learning: a systematic review and implementation guidelines}.
\newblock \bibinfo{journal}{\emph{NPJ digital medicine}} \bibinfo{volume}{3}, \bibinfo{number}{1} (\bibinfo{year}{2020}), \bibinfo{pages}{136}.
\newblock


\bibitem[Johnson et~al\mbox{.}(2023)]%
        {mimic4}
\bibfield{author}{\bibinfo{person}{Alistair~EW Johnson}, \bibinfo{person}{Lucas Bulgarelli}, \bibinfo{person}{Lu Shen}, \bibinfo{person}{Alvin Gayles}, \bibinfo{person}{Ayad Shammout}, \bibinfo{person}{Steven Horng}, \bibinfo{person}{Tom~J Pollard}, \bibinfo{person}{Sicheng Hao}, \bibinfo{person}{Benjamin Moody}, \bibinfo{person}{Brian Gow}, {et~al\mbox{.}}} \bibinfo{year}{2023}\natexlab{}.
\newblock \showarticletitle{{MIMIC-IV}, a freely accessible electronic health record dataset}.
\newblock \bibinfo{journal}{\emph{Scientific data}} \bibinfo{volume}{10}, \bibinfo{number}{1} (\bibinfo{year}{2023}), \bibinfo{pages}{1}.
\newblock


\bibitem[Johnson et~al\mbox{.}(2016)]%
        {johnson2016mimic}
\bibfield{author}{\bibinfo{person}{Alistair~EW Johnson}, \bibinfo{person}{Tom~J Pollard}, \bibinfo{person}{Lu Shen}, \bibinfo{person}{Li-wei~H Lehman}, \bibinfo{person}{Mengling Feng}, \bibinfo{person}{Mohammad Ghassemi}, \bibinfo{person}{Benjamin Moody}, \bibinfo{person}{Peter Szolovits}, \bibinfo{person}{Leo Anthony~Celi}, {and} \bibinfo{person}{Roger~G Mark}.} \bibinfo{year}{2016}\natexlab{}.
\newblock \showarticletitle{{MIMIC}-III, a freely accessible critical care database}.
\newblock \bibinfo{journal}{\emph{Scientific data}} \bibinfo{volume}{3}, \bibinfo{number}{1} (\bibinfo{year}{2016}), \bibinfo{pages}{1--9}.
\newblock


\bibitem[Kolosnjaji et~al\mbox{.}(2016)]%
        {system_call_sequences}
\bibfield{author}{\bibinfo{person}{Bojan Kolosnjaji}, \bibinfo{person}{Apostolis Zarras}, \bibinfo{person}{George Webster}, {and} \bibinfo{person}{Claudia Eckert}.} \bibinfo{year}{2016}\natexlab{}.
\newblock \showarticletitle{Deep learning for classification of malware system call sequences}. In \bibinfo{booktitle}{\emph{AI 2016: Advances in Artificial Intelligence: 29th Australasian Joint Conference}}. \bibinfo{pages}{137--149}.
\newblock


\bibitem[Leskovec and Krevl(2014)]%
        {leskovec2014snap}
\bibfield{author}{\bibinfo{person}{Jure Leskovec} {and} \bibinfo{person}{Andrej Krevl}.} \bibinfo{year}{2014}\natexlab{}.
\newblock \bibinfo{title}{SNAP Datasets: Stanford large network dataset collection}.
\newblock


\bibitem[Liu et~al\mbox{.}(2012)]%
        {10.1145/2124295.2124353}
\bibfield{author}{\bibinfo{person}{Yandong Liu}, \bibinfo{person}{Sandeep Pandey}, \bibinfo{person}{Deepak Agarwal}, {and} \bibinfo{person}{Vanja Josifovski}.} \bibinfo{year}{2012}\natexlab{}.
\newblock \showarticletitle{Finding the right consumer: optimizing for conversion in display advertising campaigns}. In \bibinfo{booktitle}{\emph{Proceedings of the Fifth ACM International Conference on Web Search and Data Mining}} (Seattle, Washington, USA) \emph{(\bibinfo{series}{WSDM '12})}. \bibinfo{publisher}{Association for Computing Machinery}, \bibinfo{address}{New York, NY, USA}, \bibinfo{pages}{473–482}.
\newblock
\showISBNx{9781450307475}
\href{https://doi.org/10.1145/2124295.2124353}{doi:\nolinkurl{10.1145/2124295.2124353}}


\bibitem[Luo et~al\mbox{.}(2014)]%
        {luo2014you}
\bibfield{author}{\bibinfo{person}{Dixin Luo}, \bibinfo{person}{Hongteng Xu}, \bibinfo{person}{Hongyuan Zha}, \bibinfo{person}{Jun Du}, \bibinfo{person}{Rong Xie}, \bibinfo{person}{Xiaokang Yang}, {and} \bibinfo{person}{Wenjun Zhang}.} \bibinfo{year}{2014}\natexlab{}.
\newblock \showarticletitle{You are what you watch and when you watch: Inferring household structures from {IPTV} viewing data}.
\newblock \bibinfo{journal}{\emph{IEEE Transactions on Broadcasting}} \bibinfo{volume}{60}, \bibinfo{number}{1} (\bibinfo{year}{2014}), \bibinfo{pages}{61--72}.
\newblock


\bibitem[Mancisidor et~al\mbox{.}(2021)]%
        {mancisidor2021learning}
\bibfield{author}{\bibinfo{person}{Rogelio~A Mancisidor}, \bibinfo{person}{Michael Kampffmeyer}, \bibinfo{person}{Kjersti Aas}, {and} \bibinfo{person}{Robert Jenssen}.} \bibinfo{year}{2021}\natexlab{}.
\newblock \showarticletitle{Learning latent representations of bank customers with the variational autoencoder}.
\newblock \bibinfo{journal}{\emph{Expert Systems with Applications}}  \bibinfo{volume}{164} (\bibinfo{date}{Feb.} \bibinfo{year}{2021}), \bibinfo{pages}{114020}.
\newblock
\href{https://doi.org/10.1016/j.eswa.2020.114020}{doi:\nolinkurl{10.1016/j.eswa.2020.114020}}


\bibitem[McDermott et~al\mbox{.}(2024)]%
        {mcdermott2024eventstreamgpt}
\bibfield{author}{\bibinfo{person}{Matthew McDermott}, \bibinfo{person}{Bret Nestor}, \bibinfo{person}{Peniel Argaw}, {and} \bibinfo{person}{Isaac~S Kohane}.} \bibinfo{year}{2024}\natexlab{}.
\newblock \showarticletitle{Event Stream GPT: a data pre-processing and modeling library for generative, pre-trained transformers over continuous-time sequences of complex events}.
\newblock \bibinfo{journal}{\emph{Advances in Neural Information Processing Systems}}  \bibinfo{volume}{36} (\bibinfo{year}{2024}).
\newblock


\bibitem[Mei and Eisner(2017)]%
        {mei2017neural}
\bibfield{author}{\bibinfo{person}{Hongyuan Mei} {and} \bibinfo{person}{Jason Eisner}.} \bibinfo{year}{2017}\natexlab{}.
\newblock \showarticletitle{The neural {H}awkes process: A neurally self-modulating multivariate point process}. In \bibinfo{booktitle}{\emph{Advances in neural information processing systems}}, Vol.~\bibinfo{volume}{30}. \bibinfo{address}{Long Beach}.
\newblock
\href{https://doi.org/10.48550/arXiv.1612.09328}{doi:\nolinkurl{10.48550/arXiv.1612.09328}}


\bibitem[Meta(2024)]%
        {meta2024llama}
\bibfield{author}{\bibinfo{person}{AI Meta}.} \bibinfo{year}{2024}\natexlab{}.
\newblock \showarticletitle{Llama 3.2: Revolutionizing edge AI and vision with open, customizable models}.
\newblock \bibinfo{journal}{\emph{Meta AI Blog. Retrieved December}}  \bibinfo{volume}{20} (\bibinfo{year}{2024}), \bibinfo{pages}{2024}.
\newblock


\bibitem[Moro et~al\mbox{.}(2014)]%
        {bank_telemarketing}
\bibfield{author}{\bibinfo{person}{Sérgio Moro}, \bibinfo{person}{Paulo Cortez}, {and} \bibinfo{person}{Paulo Rita}.} \bibinfo{year}{2014}\natexlab{}.
\newblock \showarticletitle{A data-driven approach to predict the success of bank telemarketing}. In \bibinfo{booktitle}{\emph{Decision Support Systems, 62}}. \bibinfo{pages}{22--31}.
\newblock


\bibitem[Osin et~al\mbox{.}(2024)]%
        {osin2024ebeseasybenchmarkingevent}
\bibfield{author}{\bibinfo{person}{Dmitry Osin}, \bibinfo{person}{Igor Udovichenko}, \bibinfo{person}{Viktor Moskvoretskii}, \bibinfo{person}{Egor Shvetsov}, {and} \bibinfo{person}{Evgeny Burnaev}.} \bibinfo{year}{2024}\natexlab{}.
\newblock \bibinfo{title}{EBES: Easy Benchmarking for Event Sequences}.
\newblock
\showeprint[arxiv]{2410.03399}~[cs.LG]
\urldef\tempurl%
\url{https://arxiv.org/abs/2410.03399}
\showURL{%
\tempurl}


\bibitem[Padhi et~al\mbox{.}(2021)]%
        {padhi2021tabular}
\bibfield{author}{\bibinfo{person}{Inkit Padhi}, \bibinfo{person}{Yair Schiff}, \bibinfo{person}{Igor Melnyk}, \bibinfo{person}{Mattia Rigotti}, \bibinfo{person}{Youssef Mroueh}, \bibinfo{person}{Pierre Dognin}, \bibinfo{person}{Jerret Ross}, \bibinfo{person}{Ravi Nair}, {and} \bibinfo{person}{Erik Altman}.} \bibinfo{year}{2021}\natexlab{}.
\newblock \showarticletitle{Tabular transformers for modeling multivariate time series}. In \bibinfo{booktitle}{\emph{Proceedings of International Conference on Acoustics, Speech and Signal Processing (ICASSP)}}. IEEE, \bibinfo{pages}{3565--3569}.
\newblock


\bibitem[{Pavlova}(2024)]%
        {essay101228}
\bibfield{author}{\bibinfo{person}{Miglena {Pavlova}}.} \bibinfo{year}{2024}\natexlab{}.
\newblock \bibinfo{title}{Predictive Modelling of Customer Response to Marketing Campaigns}.
\newblock
\urldef\tempurl%
\url{http://essay.utwente.nl/101228/}
\showURL{%
\tempurl}


\bibitem[Phan et~al\mbox{.}(2017)]%
        {10.1145/3077136.3080682}
\bibfield{author}{\bibinfo{person}{Minh~C. Phan}, \bibinfo{person}{Aixin Sun}, {and} \bibinfo{person}{Yi Tay}.} \bibinfo{year}{2017}\natexlab{}.
\newblock \showarticletitle{Cross-Device User Linking: URL, Session, Visiting Time, and Device-log Embedding}. In \bibinfo{booktitle}{\emph{Proceedings of the 40th International ACM SIGIR Conference on Research and Development in Information Retrieval}} (Shinjuku, Tokyo, Japan) \emph{(\bibinfo{series}{SIGIR '17})}. \bibinfo{publisher}{Association for Computing Machinery}, \bibinfo{address}{New York, NY, USA}, \bibinfo{pages}{933–936}.
\newblock
\showISBNx{9781450350228}
\href{https://doi.org/10.1145/3077136.3080682}{doi:\nolinkurl{10.1145/3077136.3080682}}


\bibitem[Radford et~al\mbox{.}(2021)]%
        {radford2021learningtransferablevisualmodels}
\bibfield{author}{\bibinfo{person}{Alec Radford}, \bibinfo{person}{Jong~Wook Kim}, \bibinfo{person}{Chris Hallacy}, \bibinfo{person}{Aditya Ramesh}, \bibinfo{person}{Gabriel Goh}, \bibinfo{person}{Sandhini Agarwal}, \bibinfo{person}{Girish Sastry}, \bibinfo{person}{Amanda Askell}, \bibinfo{person}{Pamela Mishkin}, \bibinfo{person}{Jack Clark}, \bibinfo{person}{Gretchen Krueger}, {and} \bibinfo{person}{Ilya Sutskever}.} \bibinfo{year}{2021}\natexlab{}.
\newblock \bibinfo{title}{Learning Transferable Visual Models From Natural Language Supervision}.
\newblock
\showeprint[arxiv]{2103.00020}~[cs.CV]
\urldef\tempurl%
\url{https://arxiv.org/abs/2103.00020}
\showURL{%
\tempurl}


\bibitem[Rosset et~al\mbox{.}(2001)]%
        {10.1145/502512.502581}
\bibfield{author}{\bibinfo{person}{Saharon Rosset}, \bibinfo{person}{Einat Neumann}, \bibinfo{person}{Uri Eick}, \bibinfo{person}{Nurit Vatnik}, {and} \bibinfo{person}{Izhak Idan}.} \bibinfo{year}{2001}\natexlab{}.
\newblock \showarticletitle{Evaluation of prediction models for marketing campaigns}. In \bibinfo{booktitle}{\emph{Proceedings of the Seventh ACM SIGKDD International Conference on Knowledge Discovery and Data Mining}} (San Francisco, California) \emph{(\bibinfo{series}{KDD '01})}. \bibinfo{publisher}{Association for Computing Machinery}, \bibinfo{address}{New York, NY, USA}, \bibinfo{pages}{456–461}.
\newblock
\showISBNx{158113391X}
\href{https://doi.org/10.1145/502512.502581}{doi:\nolinkurl{10.1145/502512.502581}}


\bibitem[Ruan et~al\mbox{.}(2024)]%
        {ruan2024language}
\bibfield{author}{\bibinfo{person}{Yucheng Ruan}, \bibinfo{person}{Xiang Lan}, \bibinfo{person}{Jingying Ma}, \bibinfo{person}{Yizhi Dong}, \bibinfo{person}{Kai He}, {and} \bibinfo{person}{Mengling Feng}.} \bibinfo{year}{2024}\natexlab{}.
\newblock \showarticletitle{Language modeling on tabular data: A survey of foundations, techniques and evolution}.
\newblock \bibinfo{journal}{\emph{arXiv preprint arXiv:2408.10548}} (\bibinfo{year}{2024}).
\newblock


\bibitem[Skalski et~al\mbox{.}(2023)]%
        {Skalski2023TowardsAF}
\bibfield{author}{\bibinfo{person}{Piotr Skalski}, \bibinfo{person}{David Sutton}, \bibinfo{person}{Stuart Burrell}, \bibinfo{person}{Iker Perez}, {and} \bibinfo{person}{Jason Wong}.} \bibinfo{year}{2023}\natexlab{}.
\newblock \showarticletitle{Towards a Foundation Purchasing Model: Pretrained Generative Autoregression on Transaction Sequences}.
\newblock \bibinfo{journal}{\emph{Proceedings of the Fourth ACM International Conference on AI in Finance}} (\bibinfo{year}{2023}).
\newblock
\urldef\tempurl%
\url{https://api.semanticscholar.org/CorpusID:265453725}
\showURL{%
\tempurl}


\bibitem[Taghibakhshi et~al\mbox{.}(2023)]%
        {taghibakhshi2023hierarchical}
\bibfield{author}{\bibinfo{person}{Ali Taghibakhshi}, \bibinfo{person}{Mingyuan Ma}, \bibinfo{person}{Ashwath Aithal}, \bibinfo{person}{Onur Yilmaz}, \bibinfo{person}{Haggai Maron}, {and} \bibinfo{person}{Matthew West}.} \bibinfo{year}{2023}\natexlab{}.
\newblock \showarticletitle{Hierarchical graph neural network with cross-attention for cross-device user matching}. In \bibinfo{booktitle}{\emph{International Conference on Big Data Analytics and Knowledge Discovery}}. Springer, \bibinfo{pages}{303--315}.
\newblock


\bibitem[Tsitsulin et~al\mbox{.}(2023)]%
        {tsitsulin2023unsupervised}
\bibfield{author}{\bibinfo{person}{Anton Tsitsulin}, \bibinfo{person}{Marina Munkhoeva}, {and} \bibinfo{person}{Bryan Perozzi}.} \bibinfo{year}{2023}\natexlab{}.
\newblock \bibinfo{title}{Unsupervised Embedding Quality Evaluation}.
\newblock
\showeprint[arxiv]{2305.16562}~[cs.LG]


\bibitem[Udovichenko et~al\mbox{.}(2024)]%
        {Udovichenko_2024}
\bibfield{author}{\bibinfo{person}{Igor Udovichenko}, \bibinfo{person}{Egor Shvetsov}, \bibinfo{person}{Denis Divitsky}, \bibinfo{person}{Dmitry Osin}, \bibinfo{person}{Ilya Trofimov}, \bibinfo{person}{Ivan Sukharev}, \bibinfo{person}{Anatoliy Glushenko}, \bibinfo{person}{Dmitry Berestnev}, {and} \bibinfo{person}{Evgeny Burnaev}.} \bibinfo{year}{2024}\natexlab{}.
\newblock \showarticletitle{SeqNAS: Neural Architecture Search for Event Sequence Classification}.
\newblock \bibinfo{journal}{\emph{IEEE Access}}  \bibinfo{volume}{12} (\bibinfo{year}{2024}), \bibinfo{pages}{3898–3909}.
\newblock
\showISSN{2169-3536}
\href{https://doi.org/10.1109/access.2024.3349497}{doi:\nolinkurl{10.1109/access.2024.3349497}}


\bibitem[Vakhrushev et~al\mbox{.}(2021)]%
        {vakhrushev2021lightautoml}
\bibfield{author}{\bibinfo{person}{Anton Vakhrushev}, \bibinfo{person}{Alexander Ryzhkov}, \bibinfo{person}{Maxim Savchenko}, \bibinfo{person}{Dmitry Simakov}, \bibinfo{person}{Rinchin Damdinov}, {and} \bibinfo{person}{Alexander Tuzhilin}.} \bibinfo{year}{2021}\natexlab{}.
\newblock \showarticletitle{Lightautoml: Automl solution for a large financial services ecosystem}.
\newblock \bibinfo{journal}{\emph{arXiv preprint arXiv:2109.01528}} (\bibinfo{year}{2021}).
\newblock


\bibitem[Vaswani et~al\mbox{.}(2017)]%
        {Vaswani2017AttentionIA}
\bibfield{author}{\bibinfo{person}{Ashish Vaswani}, \bibinfo{person}{Noam~M. Shazeer}, \bibinfo{person}{Niki Parmar}, \bibinfo{person}{Jakob Uszkoreit}, \bibinfo{person}{Llion Jones}, \bibinfo{person}{Aidan~N. Gomez}, \bibinfo{person}{Lukasz Kaiser}, {and} \bibinfo{person}{Illia Polosukhin}.} \bibinfo{year}{2017}\natexlab{}.
\newblock \showarticletitle{Attention is All you Need}. In \bibinfo{booktitle}{\emph{Neural Information Processing Systems}}.
\newblock
\urldef\tempurl%
\url{https://api.semanticscholar.org/CorpusID:13756489}
\showURL{%
\tempurl}


\bibitem[Verma et~al\mbox{.}(2020)]%
        {Verma2020GeoHashTB}
\bibfield{author}{\bibinfo{person}{Jai~Prakash Verma}, \bibinfo{person}{Sapan~H. Mankad}, {and} \bibinfo{person}{Sanjay Garg}.} \bibinfo{year}{2020}\natexlab{}.
\newblock \showarticletitle{GeoHash tag based mobility detection and prediction for traffic management}.
\newblock \bibinfo{journal}{\emph{SN Applied Sciences}}  \bibinfo{volume}{2} (\bibinfo{year}{2020}).
\newblock
\urldef\tempurl%
\url{https://api.semanticscholar.org/CorpusID:225509721}
\showURL{%
\tempurl}


\bibitem[Wang and Xiao(2022)]%
        {wang2022deep}
\bibfield{author}{\bibinfo{person}{Chongren Wang} {and} \bibinfo{person}{Zhuoyi Xiao}.} \bibinfo{year}{2022}\natexlab{}.
\newblock \showarticletitle{A Deep Learning Approach for Credit Scoring Using Feature Embedded Transformer}.
\newblock \bibinfo{journal}{\emph{Applied Sciences (Basel)}} \bibinfo{volume}{12}, \bibinfo{number}{21} (\bibinfo{date}{Oct.} \bibinfo{year}{2022}), \bibinfo{pages}{10995}.
\newblock
\href{https://doi.org/10.3390/app122110995}{doi:\nolinkurl{10.3390/app122110995}}


\bibitem[Weiss and Hirsh(1998)]%
        {Rare_Events}
\bibfield{author}{\bibinfo{person}{Gary~M. Weiss} {and} \bibinfo{person}{Haym Hirsh}.} \bibinfo{year}{1998}\natexlab{}.
\newblock \showarticletitle{Learning to Predict Rare Events in Event Sequences}. In \bibinfo{booktitle}{\emph{KDD}}. \bibinfo{pages}{359--363}.
\newblock


\bibitem[Xu and Zha(2017)]%
        {xu2017dirichlet}
\bibfield{author}{\bibinfo{person}{Hongteng Xu} {and} \bibinfo{person}{Hongyuan Zha}.} \bibinfo{year}{2017}\natexlab{}.
\newblock \showarticletitle{A {D}irichlet mixture model of {H}awkes processes for event sequence clustering}.
\newblock \bibinfo{journal}{\emph{NeurIPS}}  \bibinfo{volume}{30} (\bibinfo{year}{2017}).
\newblock


\bibitem[Xu et~al\mbox{.}(2023)]%
        {xu2023multimodallearningtransformerssurvey}
\bibfield{author}{\bibinfo{person}{Peng Xu}, \bibinfo{person}{Xiatian Zhu}, {and} \bibinfo{person}{David~A Clifton}.} \bibinfo{year}{2023}\natexlab{}.
\newblock \showarticletitle{Multimodal learning with transformers: A survey}.
\newblock \bibinfo{journal}{\emph{IEEE Transactions on Pattern Analysis and Machine Intelligence}} \bibinfo{volume}{45}, \bibinfo{number}{10} (\bibinfo{year}{2023}), \bibinfo{pages}{12113--12132}.
\newblock


\bibitem[Yang et~al\mbox{.}(2024)]%
        {yang2024qwen2}
\bibfield{author}{\bibinfo{person}{An Yang}, \bibinfo{person}{Baosong Yang}, \bibinfo{person}{Beichen Zhang}, \bibinfo{person}{Binyuan Hui}, \bibinfo{person}{Bo Zheng}, \bibinfo{person}{Bowen Yu}, \bibinfo{person}{Chengyuan Li}, \bibinfo{person}{Dayiheng Liu}, \bibinfo{person}{Fei Huang}, \bibinfo{person}{Haoran Wei}, {et~al\mbox{.}}} \bibinfo{year}{2024}\natexlab{}.
\newblock \showarticletitle{Qwen2. 5 technical report}.
\newblock \bibinfo{journal}{\emph{arXiv preprint arXiv:2412.15115}} (\bibinfo{year}{2024}).
\newblock


\bibitem[Yeshchenko and Mendling(2022)]%
        {yeshchenko2022surveyapproacheseventsequence}
\bibfield{author}{\bibinfo{person}{Anton Yeshchenko} {and} \bibinfo{person}{Jan Mendling}.} \bibinfo{year}{2022}\natexlab{}.
\newblock \bibinfo{title}{A Survey of Approaches for Event Sequence Analysis and Visualization using the ESeVis Framework}.
\newblock
\showeprint[arxiv]{2202.07941}~[cs.HC]
\urldef\tempurl%
\url{https://arxiv.org/abs/2202.07941}
\showURL{%
\tempurl}


\bibitem[Zhang et~al\mbox{.}(2023)]%
        {zhang2023fata}
\bibfield{author}{\bibinfo{person}{Dongyu Zhang}, \bibinfo{person}{Liang Wang}, \bibinfo{person}{Xin Dai}, \bibinfo{person}{Shubham Jain}, \bibinfo{person}{Junpeng Wang}, \bibinfo{person}{Yujie Fan}, \bibinfo{person}{Chin-Chia~Michael Yeh}, \bibinfo{person}{Yan Zheng}, \bibinfo{person}{Zhongfang Zhuang}, {and} \bibinfo{person}{Wei Zhang}.} \bibinfo{year}{2023}\natexlab{}.
\newblock \showarticletitle{Fata-trans: Field and time-aware transformer for sequential tabular data}. In \bibinfo{booktitle}{\emph{Proceedings of the 32nd ACM International Conference on Information and Knowledge Management}}. \bibinfo{pages}{3247--3256}.
\newblock


\bibitem[Zhang and Yan(2023)]%
        {zhang2023crossformer}
\bibfield{author}{\bibinfo{person}{Yunhao Zhang} {and} \bibinfo{person}{Junchi Yan}.} \bibinfo{year}{2023}\natexlab{}.
\newblock \showarticletitle{Crossformer: Transformer Utilizing Cross-Dimension Dependency for Multivariate Time Series Forecasting}. In \bibinfo{booktitle}{\emph{The Eleventh International Conference on Learning Representations}}.
\newblock
\urldef\tempurl%
\url{https://openreview.net/forum?id=vSVLM2j9eie}
\showURL{%
\tempurl}


\bibitem[Zhao et~al\mbox{.}(2015)]%
        {zhao2015seismic}
\bibfield{author}{\bibinfo{person}{Qingyuan Zhao}, \bibinfo{person}{Murat~A Erdogdu}, \bibinfo{person}{Hera~Y He}, \bibinfo{person}{Anand Rajaraman}, {and} \bibinfo{person}{Jure Leskovec}.} \bibinfo{year}{2015}\natexlab{}.
\newblock \showarticletitle{Seismic: A self-exciting point process model for predicting tweet popularity}. In \bibinfo{booktitle}{\emph{ACM SIGKDD international conference on knowledge discovery and data mining}}. \bibinfo{pages}{1513--1522}.
\newblock


\bibitem[Zhuzhel et~al\mbox{.}(2023)]%
        {Zhuzhel2023ContinuoustimeCM}
\bibfield{author}{\bibinfo{person}{Vladislav Zhuzhel}, \bibinfo{person}{Vsevolod Grabar}, \bibinfo{person}{Galina Boeva}, \bibinfo{person}{Artem Zabolotnyi}, \bibinfo{person}{Alexander Stepikin}, \bibinfo{person}{Vladimir~V. Zholobov}, \bibinfo{person}{Maria Ivanova}, \bibinfo{person}{Mikhail Orlov}, \bibinfo{person}{Ivan~A Kireev}, \bibinfo{person}{Evgeny~V. Burnaev}, \bibinfo{person}{Rodrigo Rivera-Castro}, {and} \bibinfo{person}{Alexey Zaytsev}.} \bibinfo{year}{2023}\natexlab{}.
\newblock \showarticletitle{Continuous-time convolutions model of event sequences}.
\newblock \bibinfo{journal}{\emph{ArXiv}}  \bibinfo{volume}{abs/2302.06247} (\bibinfo{year}{2023}).
\newblock
\urldef\tempurl%
\url{https://api.semanticscholar.org/CorpusID:256827392}
\showURL{%
\tempurl}


\bibitem[Zong et~al\mbox{.}(2023)]%
        {zong2023selfsupervised}
\bibfield{author}{\bibinfo{person}{Yongshuo Zong}, \bibinfo{person}{Oisin~Mac Aodha}, {and} \bibinfo{person}{Timothy Hospedales}.} \bibinfo{year}{2023}\natexlab{}.
\newblock \showarticletitle{Self-Supervised Multimodal Learning: A Survey}.
\newblock  (\bibinfo{year}{2023}).
\newblock
\showeprint[arxiv]{2304.01008}~[cs.LG]


\end{thebibliography}

\clearpage
\appendix

\section{Dataset Statistics and data samples}\label{sec:appendix_stat}

\begin{figure}[h]
  \centering
  \includegraphics[width=0.9\columnwidth]{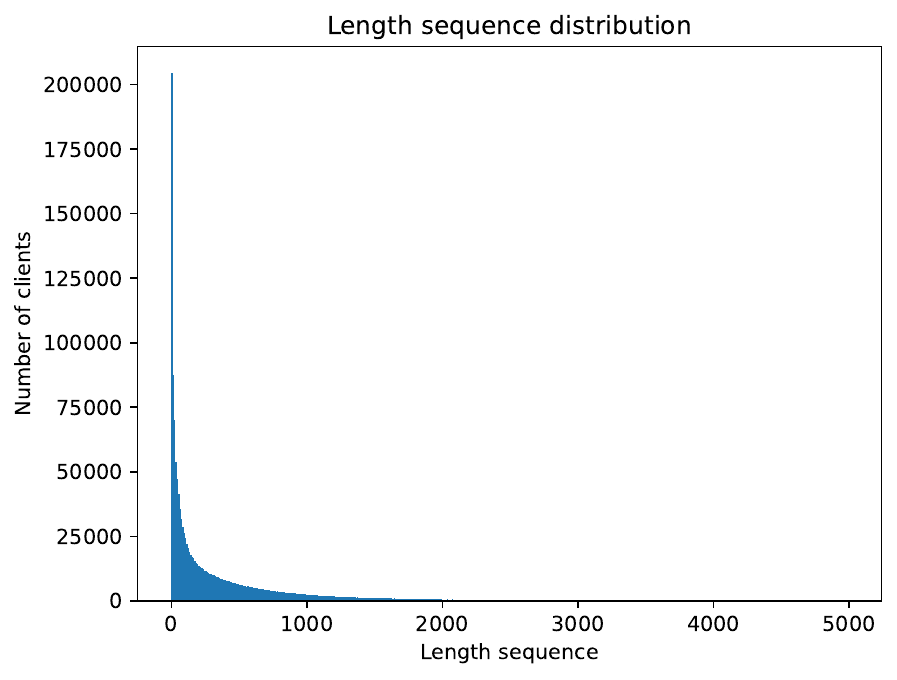}
 \caption{The histogram of the number of clients with a certain length of transaction history}
  \Description{The histogram of the number of clients with a certain length of transaction history.}
 \label{fig:histo_transactions}
\end{figure}

\begin{figure}[h]
  \centering
  \includegraphics[width=\linewidth]{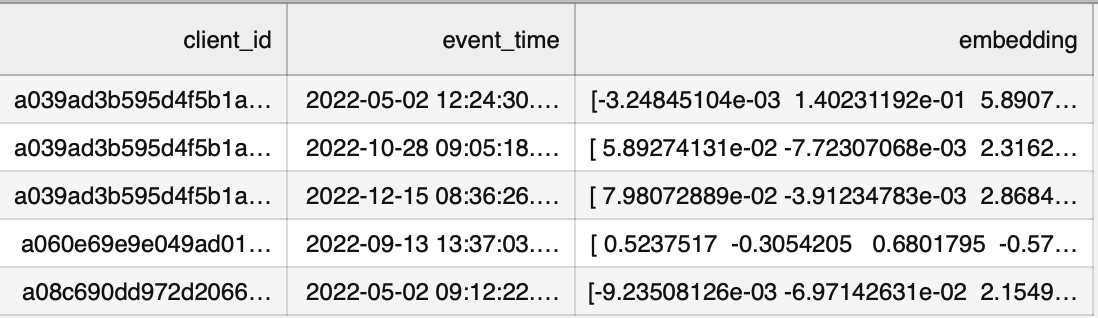}
 \caption{Sample data of dialogues modality.}
  \Description{Sample data of dialogues modality.}
 \label{fig:sample_dialogues}
\end{figure}

\begin{figure}[h]
  \centering
  \includegraphics[width=0.9\linewidth]{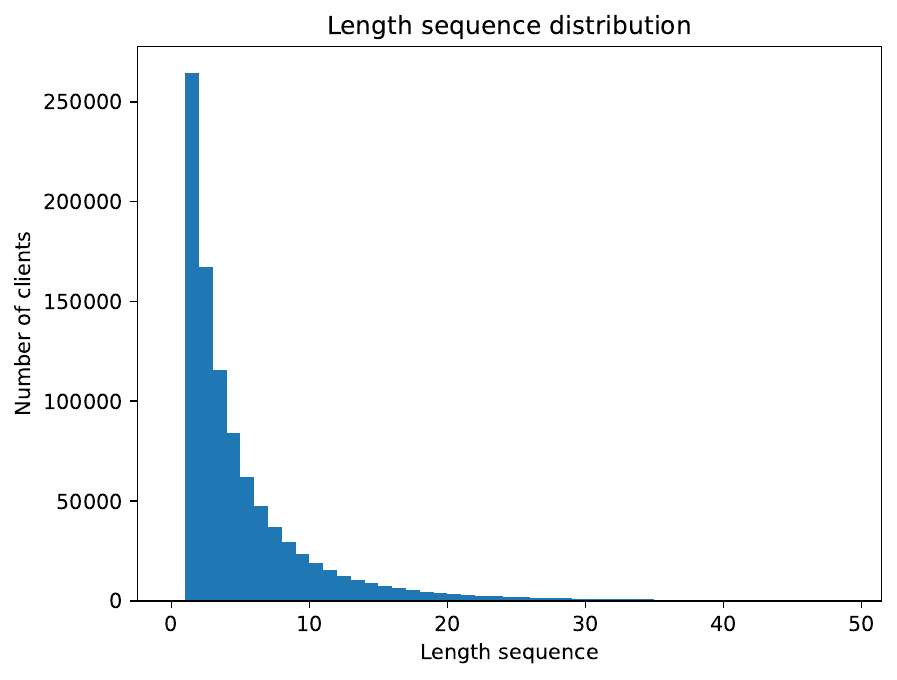}
 \caption{The histogram of the number of clients with a certain number of dialogues.}
  \Description{The histogram of the number of clients with a certain number of dialogues.}
 \label{fig:histo_dialogues}
\end{figure}

\begin{figure}[h]
  \centering
  \includegraphics[width=\linewidth]{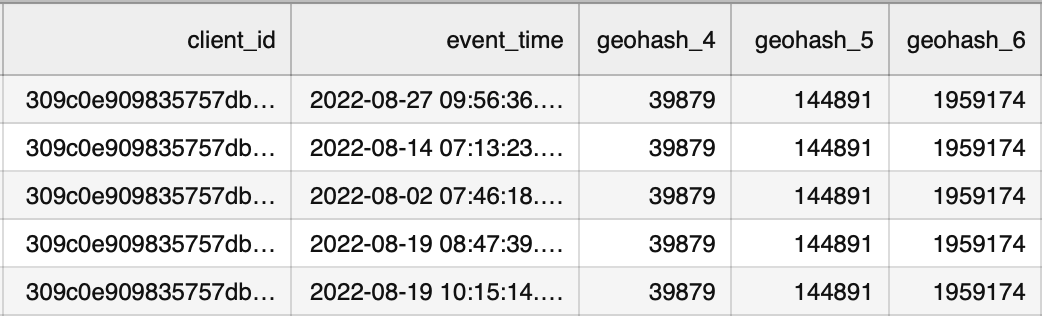}
 \caption{Sample data of geostream modality.}
  \Description{Sample data of geostream modality.}
 \label{fig:sample_geo}
\end{figure}

\begin{figure}[h]
  \centering
  \includegraphics[width=0.9\linewidth]{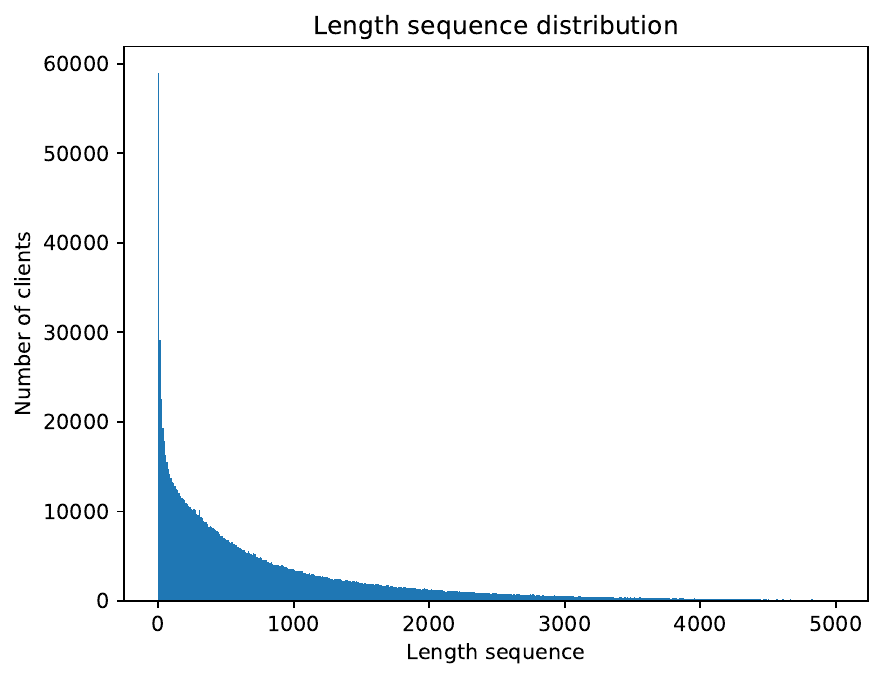}
 \caption{The histogram of the number of clients with a certain length of geostream.}
  \Description{The histogram of the number of clients with a certain length of geostream.}
 \label{fig:histo_geostream}
\end{figure}

\begin{figure}[h]
  \centering
  \includegraphics[width=\linewidth]{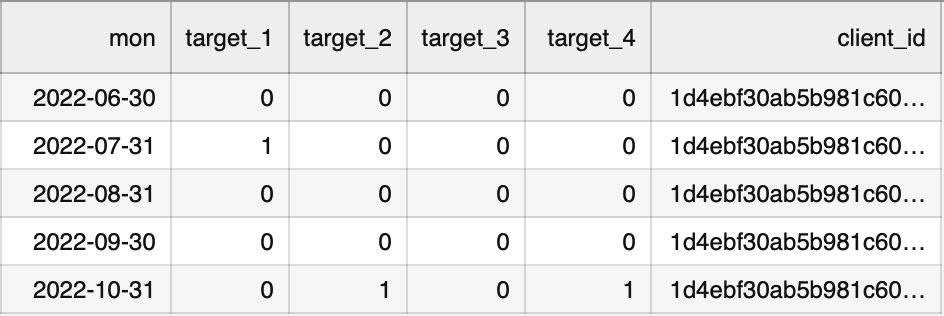}
 \caption{Example data of client's purchases.}
  \Description{Example data of client's purchases.}
 \label{fig:sample_purchases}
\end{figure}

In this section, we present data samples from various modalities and a histogram of sequence lengths. Table~\ref{tab:transaction_attrs} provides a list of attributes for the transactional modalities, while Figure~\ref{fig:histo_transactions} illustrates the distribution of sequence lengths for these modalities. Additionally, Figure~\ref{fig:sample_geo} shows a sample of geographical data, and Figure~\ref{fig:histo_geostream} displays the distribution of sequence lengths for geostream modalities. For dialogue data, a sample is presented in Figure~\ref{fig:sample_dialogues}, with the distribution of sequence lengths for dialogue events shown in Figure~\ref{fig:histo_dialogues}. Figure~\ref{fig:sample_purchases} demonstrates a sample of data for product purchases. Here, we observe that geographical and transactional modalities have long tails in the histogram of distributions. 

\begin{table}
 \caption{Transaction features}
 \label{tab:transaction_attrs}
 \centering
 \begin{tabular}{lll}
 \toprule
 Features & Number of categories & Description \\
 \midrule
 amount & - & transaction amount \\
 event time & - & transaction time \\
 currency & 15 & transfer currency \\
 event type & 56 & transaction type\\
 event subtype & 62 & transaction subtype \\
 src type11 & 101 & sender field type 1\\
 src type12 & 536 & sender field subtype 1\\
 dst type11 & 123 & receiver field type 1\\
 dst type12 & 649 & receiver field subtype 1\\
 src type21 & 40131 & sender field type 2\\
 src type22 & 87 & sender field subtype 2\\
 src type31 & 2389 & sender field type 3\\
 src type32 & 88 & sender field subtype 3\\ 
 \bottomrule
 \end{tabular}
\end{table}

\begin{figure}[h]
  \centering
  \includegraphics[height=180pt, width=180pt]{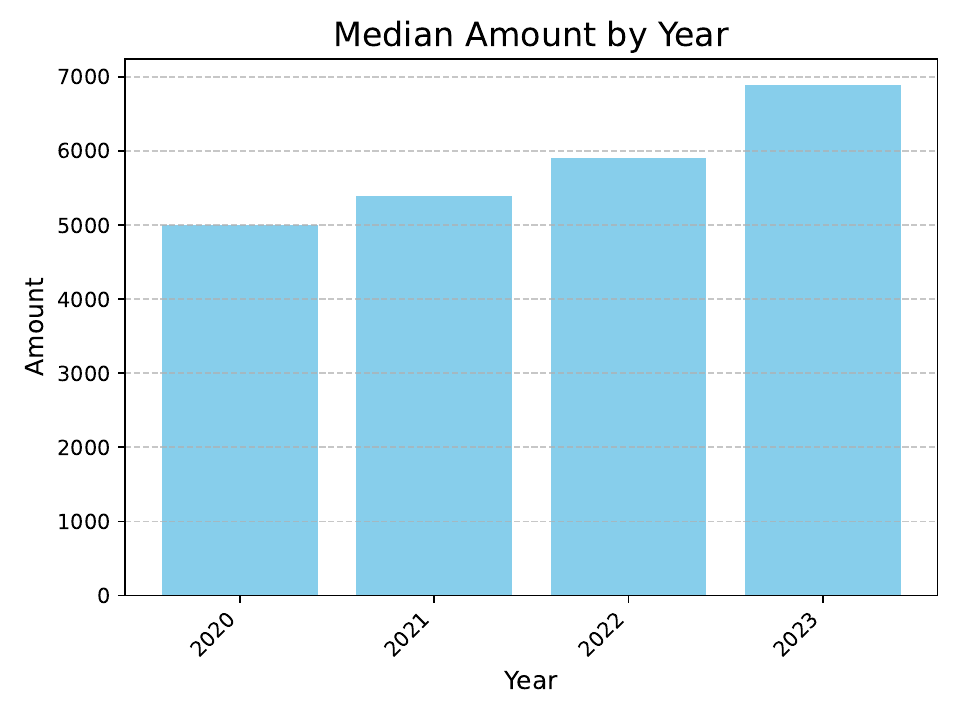}
 \caption{\textbf{Median amount per client for private transaction data.} The data reveals a trend of increasing median amounts over four years.}
  \Description{Median amount per client for private transaction data.}
 \label{fig:median_amount}
\end{figure}
\begin{figure}[h]
  \centering
  \includegraphics[height=180pt, width=180pt]{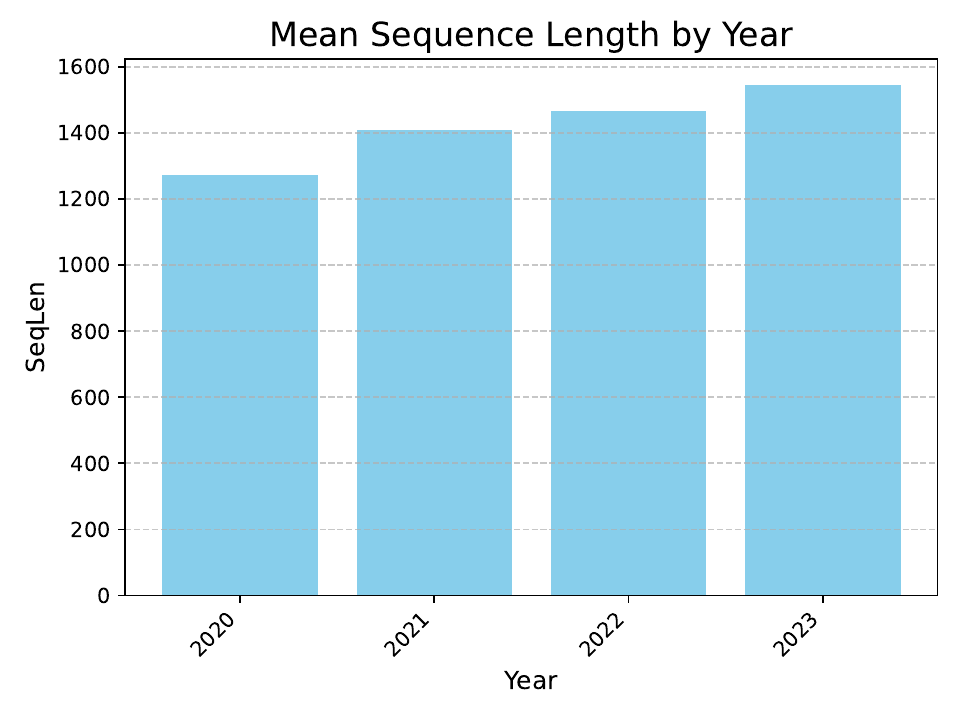}
 \caption{\textbf{Mean sequence length per clients for private transaction data.} The mean sequence length has slightly increased over the past four years} 
  \Description{Mean sequence length per client for private transaction data.}
 \label{fig:mean_seqlen}
\end{figure}

\section{Detailed experimental results}\label{sec:appendix_expers}

\subsection{Multimodal matching}
For the multimodal matching task Table~\ref{table:matching}, it is observed that the dialogue modality exhibits poor compatibility when combined with other modalities.

\begin{table}
 \caption{MBD: Detailed multimodal matching results}
 \label{table:matching}
 \centering
 \begin{tabular}{l|*{3}{c}}
 \toprule
&Recall@1&Recall@50&Recall@100 \\
\midrule
Trx2Geo&0.006 $\pm$ 0.0003&0.1967 $\pm$ 0.002& 0.303 $\pm$ 0.004 \\
Geo2Trx&0.004 $\pm$ 0.0003&0.1624 $\pm$ 0.002&0.262 $\pm$ 0.004 \\
Trx2Dial&0.00004 $\pm$ 0.00001&0.0001 $\pm$ 0.00005&0.003 $\pm$ 0.0001 \\
Dial2Trx&0.00003 $\pm$ 0.00001&0.001 $\pm$ 0.00005&0.003 $\pm$ 0.0001 \\
Dial2Geo& 0.00002 $\pm$ 0.000001&0.0006 $\pm$ 0.00001&0.001 $\pm$ 0.0001 \\
Geo2Dial& 0.00002 $\pm$ 0.000001&0.0005 $\pm$ 0.00001&0.001 $\pm$ 0.0001 \\
\bottomrule
\end{tabular}
\end{table}

\subsection{Handling class imbalance}

We conducted experiments on the MBD dataset, focusing on transaction modalities, and explored various techniques, including random undersampling, oversampling, and class balancing within gradient boosting for CoLES embeddings. Additionally, stratified batching was applied to better align the target distribution in Supervised RNN. The results of these experiments are summarized in Table~\ref{table:label_imbalance_results}.

\begin{table}
\centering
\caption{Performance Comparison for Label Imbalance on MBD (Transactions)}
\label{table:label_imbalance_results}
\resizebox{\columnwidth}{!}{%
\begin{tabular}{lcc}
\toprule
\textbf{Method}                                & \textbf{Description}                  & \textbf{ROC-AUC} \\
\midrule
\textbf{Supervised RNN}                        & Baseline                              & 0.819            \\
\textbf{Supervised RNN (stratified batching)}  & Balanced target distribution          & 0.819            \\
\textbf{CoLES}                                 & Baseline embeddings                   & 0.773            \\
\textbf{CoLES (undersampling)}                 & Random undersampling applied          & 0.772            \\
\textbf{CoLES (oversampling)}                  & Random oversampling applied           & 0.774            \\
\textbf{CoLES (class balanced)}                & Class balancing in gradient boosting  & 0.774            \\ 
\bottomrule
\end{tabular}%
}
\end{table}

Here, stratified batching consistently maintains a high ROC-AUC of 0.819 for Supervised RNN, demonstrating its robustness to label imbalance. For CoLES embeddings, balancing techniques result in minimal variations in performance, with ROC-AUC values ranging from 0.772 to 0.774. This indicates that CoLES embeddings exhibit limited sensitivity to label imbalance, highlighting the potential need for more advanced balancing strategies to achieve further improvements.

\subsection{Campaining results on the private dataset and MBD}
The experimental results presented in Tables \ref{table:public_blending} and \ref{table:public_fusion} underscore the pivotal importance of modality fusion in enhancing model performance. 
Here, integrating dialogue data with transaction data leads to a significant performance boost. For instance, augmenting TrxTabGPT with DialogLast results in a 1.8\% improvement in the mean metric for blending and a 3.4\% enhancement in fusion. The most substantial performance gains are observed when all modalities are combined. Specifically, the integration of TrxTabGPT, GeoTabGPT, and DialogLast yields a 2\% improvement over unimodal transaction models and a 3.2\% improvement in late fusion, highlighting the synergistic benefits of incorporating dialogue and geographical data into transaction-based models. These findings provide robust evidence supporting the effectiveness of multimodal integration. The inclusion of dialogue and geographical data significantly boosts the performance of models centered on transaction data. Moreover, this trend observed in public datasets is consistently replicated in proprietary data, as shown in Tables \ref{table:private_blending} and \ref{table:private_fusion}.

These results also prove the low impact of our anonymization procedure. Indeed, the ranking of methods remains consistent despite differences in absolute metric values. As shown in Table~\ref{table:private_fusion}, TrxTabGPT (ROC-AUC 0.796) and TrxAggregation (ROC-AUC 0.780) achieve the best performance on the private dataset. Similarly, TrxTabGPT leads on the public dataset with an ROC-AUC of 0.802 (Table~\ref{table:public_fusion}). Incorporating geolocation and dialogue modalities further improves results, with DialogLast+TrxTabGPT+GeoTabGPT attaining the highest ROC-AUC on both datasets: 0.802 on the private dataset (Table~\ref{table:private_fusion}) and 0.808 on the public dataset (Table~\ref{table:public_fusion}). Overall, multimodal approaches utilizing TabGPT or Aggregation demonstrate superior performance, with Late Fusion consistently outperforming Blending across private and public datasets (compare results in Table~\ref{table:private_blending} and Table~\ref{table:public_blending} to those in Table~\ref{table:private_fusion} and Table~\ref{table:public_fusion}).

\begin{table*}
 \caption{Blending results on private dataset}
\label{table:private_blending}
 \centering
 \resizebox{\textwidth}{!}{
 \begin{tabular}{l|*1{c}|*{4}{c}}
 \toprule
Methods&mean & target\_1&target\_2&target\_3& target\_4 \\
\midrule
DialogLast & 0.590 $\pm$ 0.001 & 0.633 $\pm$ 0.001 & 0.604 $\pm$ 0.005 & 0.549 $\pm$ 0.002 & 0.576 $\pm$ 0.002 \\
DialogLast+GeoAggregation & 0.603 $\pm$ 0.002 & 0.632 $\pm$ 0.001 & 0.629 $\pm$ 0.007 & 0.558 $\pm$ 0.001 & 0.592 $\pm$ 0.002 \\
DialogLast+GeoCoLES & 0.632 $\pm$ 0.002 & 0.641 $\pm$ 0.001 & 0.688 $\pm$ 0.008 & 0.580 $\pm$ 0.002 & 0.619 $\pm$ 0.003 \\
DialogLast+GeoTabGPT & 0.648 $\pm$ 0.002 & 0.654 $\pm$ 0.001 & 0.715 $\pm$ 0.008 & 0.591 $\pm$ 0.003 & 0.632 $\pm$ 0.003 \\
DialogLast+GeoTabBERT & 0.628 $\pm$ 0.001 & 0.641 $\pm$ 0.002 & 0.683 $\pm$ 0.006 & 0.579 $\pm$ 0.002 & 0.610 $\pm$ 0.006 \\
DialogLast+TrxAggregation & 0.731 $\pm$ 0.001 & 0.701 $\pm$ 0.001 & 0.816 $\pm$ 0.003 & 0.673 $\pm$ 0.001 & 0.735 $\pm$ 0.002 \\
DialogLast+TrxAggregation+GeoAggregation & 0.729 $\pm$ 0.001 & 0.694 $\pm$ 0.001 & 0.812 $\pm$ 0.003 & 0.670 $\pm$ 0.001 & 0.738 $\pm$ 0.003 \\
DialogLast+TrxCoLES & 0.715 $\pm$ 0.002 & 0.692 $\pm$ 0.002 & 0.800 $\pm$ 0.005 & 0.645 $\pm$ 0.002 & 0.723 $\pm$ 0.005 \\
DialogLast+TrxCoLES+GeoCoLES & 0.720 $\pm$ 0.002 & 0.689 $\pm$ 0.003 & 0.815 $\pm$ 0.003 & 0.644 $\pm$ 0.002 & 0.734 $\pm$ 0.005 \\
DialogLast+TrxTabGPT & 0.739 $\pm$ 0.001 & 0.683 $\pm$ 0.001 & 0.820 $\pm$ 0.005 & 0.685 $\pm$ 0.001 & 0.767 $\pm$ 0.005 \\
DialogLast+TrxTabGPT+GeoTabGPT & 0.749 $\pm$ 0.002 & 0.690 $\pm$ 0.001 & 0.838 $\pm$ 0.005 & 0.687 $\pm$ 0.001 & 0.780 $\pm$ 0.005 \\
DialogLast+TrxTabBERT & 0.710 $\pm$ 0.006 & 0.686 $\pm$ 0.004 & 0.806 $\pm$ 0.004 & 0.628 $\pm$ 0.006 & 0.720 $\pm$ 0.012 \\
DialogLast+TrxTabBERT+GeoTabBERT & 0.716 $\pm$ 0.005 & 0.683 $\pm$ 0.003 & 0.821 $\pm$ 0.003 & 0.627 $\pm$ 0.006 & 0.731 $\pm$ 0.009 \\
DialogMean & 0.604 $\pm$ 0.001 & 0.636 $\pm$ 0.002 & 0.629 $\pm$ 0.001 & 0.564 $\pm$ 0.001 & 0.587 $\pm$ 0.001 \\
DialogMean+GeoAggregation & 0.613 $\pm$ 0.001 & 0.635 $\pm$ 0.002 & 0.648 $\pm$ 0.001 & 0.568 $\pm$ 0.001 & 0.601 $\pm$ 0.002 \\
DialogMean+GeoCoLES & 0.638 $\pm$ 0.002 & 0.643 $\pm$ 0.002 & 0.699 $\pm$ 0.005 & 0.585 $\pm$ 0.001 & 0.626 $\pm$ 0.001 \\
DialogMean+GeoTabGPT & 0.653 $\pm$ 0.001 & 0.656 $\pm$ 0.001 & 0.725 $\pm$ 0.005 & 0.595 $\pm$ 0.002 & 0.638 $\pm$ 0.003 \\
DialogMean+GeoTabBERT & 0.635 $\pm$ 0.002 & 0.643 $\pm$ 0.002 & 0.696 $\pm$ 0.004 & 0.584 $\pm$ 0.001 & 0.617 $\pm$ 0.006 \\
DialogMean+TrxAggregation & 0.732 $\pm$ 0.001 & 0.700 $\pm$ 0.001 & 0.816 $\pm$ 0.003 & 0.673 $\pm$ 0.001 & 0.739 $\pm$ 0.001 \\
DialogMean+TrxAggregation+GeoAggregation & 0.729 $\pm$ 0.001 & 0.694 $\pm$ 0.001 & 0.812 $\pm$ 0.004 & 0.670 $\pm$ 0.001 & 0.741 $\pm$ 0.002 \\
DialogMean+TrxCoLES & 0.715 $\pm$ 0.002 & 0.692 $\pm$ 0.002 & 0.800 $\pm$ 0.004 & 0.645 $\pm$ 0.002 & 0.725 $\pm$ 0.004 \\
DialogMean+TrxCoLES+GeoCoLES & 0.721 $\pm$ 0.002 & 0.689 $\pm$ 0.003 & 0.813 $\pm$ 0.003 & 0.644 $\pm$ 0.002 & 0.737 $\pm$ 0.004 \\
DialogMean+TrxTabGPT & 0.739 $\pm$ 0.001 & 0.682 $\pm$ 0.001 & 0.819 $\pm$ 0.005 & 0.684 $\pm$ 0.001 & 0.769 $\pm$ 0.004 \\
DialogMean+TrxTabGPT+GeoTabGPT & 0.748 $\pm$ 0.001 & 0.689 $\pm$ 0.001 & 0.837 $\pm$ 0.004 & 0.686 $\pm$ 0.001 & 0.782 $\pm$ 0.004 \\
DialogMean+TrxTabBERT & 0.711 $\pm$ 0.006 & 0.685 $\pm$ 0.004 & 0.806 $\pm$ 0.005 & 0.629 $\pm$ 0.005 & 0.726 $\pm$ 0.011 \\
DialogMean+TrxTabBERT+GeoTabBERT & 0.717 $\pm$ 0.005 & 0.682 $\pm$ 0.003 & 0.821 $\pm$ 0.004 & 0.628 $\pm$ 0.006 & 0.735 $\pm$ 0.009 \\
GeoAggregation & 0.554 $\pm$ 0.001 & 0.540 $\pm$ 0.001 & 0.584 $\pm$ 0.002 & 0.534 $\pm$ 0.001 & 0.559 $\pm$ 0.001 \\
GeoCoLES & 0.601 $\pm$ 0.004 & 0.565 $\pm$ 0.004 & 0.668 $\pm$ 0.011 & 0.571 $\pm$ 0.003 & 0.600 $\pm$ 0.003 \\
GeoTabGPT & 0.622 $\pm$ 0.001 & 0.589 $\pm$ 0.001 & 0.700 $\pm$ 0.008 & 0.586 $\pm$ 0.003 & 0.615 $\pm$ 0.004 \\
GeoTabBERT & 0.596 $\pm$ 0.002 & 0.566 $\pm$ 0.003 & 0.663 $\pm$ 0.010 & 0.570 $\pm$ 0.003 & 0.585 $\pm$ 0.007 \\
TrxAggregation & 0.783 $\pm$ 0.001 & 0.743 $\pm$ 0.001 & 0.825 $\pm$ 0.002 & 0.764 $\pm$ 0.001 & 0.801 $\pm$ 0.002 \\
TrxAggregation+GeoAggregation & 0.774 $\pm$ 0.001 & 0.733 $\pm$ 0.001 & 0.817 $\pm$ 0.003 & 0.756 $\pm$ 0.001 & 0.789 $\pm$ 0.003 \\
TrxCoLES & 0.772 $\pm$ 0.002 & 0.734 $\pm$ 0.003 & 0.813 $\pm$ 0.004 & 0.747 $\pm$ 0.002 & 0.793 $\pm$ 0.003 \\
TrxCoLES+GeoCoLES & 0.772 $\pm$ 0.002 & 0.729 $\pm$ 0.003 & 0.825 $\pm$ 0.006 & 0.740 $\pm$ 0.002 & 0.795 $\pm$ 0.003 \\
TrxTabGPT & 0.796 $\pm$ 0.000 & 0.746 $\pm$ 0.001 & 0.837 $\pm$ 0.004 & 0.778 $\pm$ 0.001 & 0.825 $\pm$ 0.004 \\
TrxTabGPT+GeoTabGPT & 0.798 $\pm$ 0.001 & 0.743 $\pm$ 0.001 & 0.850 $\pm$ 0.003 & 0.772 $\pm$ 0.001 & 0.827 $\pm$ 0.004 \\
TrxTabBERT & 0.754 $\pm$ 0.011 & 0.707 $\pm$ 0.019 & 0.815 $\pm$ 0.006 & 0.717 $\pm$ 0.012 & 0.778 $\pm$ 0.012 \\
TrxTabBERT+GeoTabBERT & 0.758 $\pm$ 0.010 & 0.707 $\pm$ 0.016 & 0.831 $\pm$ 0.006 & 0.713 $\pm$ 0.011 & 0.781 $\pm$ 0.010 \\
\bottomrule
\end{tabular}
}
\end{table*}

\begin{table*}
 \caption{Late Fusion results on private dataset}
\label{table:private_fusion}
 \centering
 \resizebox{\textwidth}{!}{
 \begin{tabular}{l|*1{c}|*{4}{c}}
 \toprule
Methods&mean & target\_1&target\_2&target\_3& target\_4 \\
\midrule
DialogLast & 0.590 $\pm$ 0.001 & 0.633 $\pm$ 0.001 & 0.604 $\pm$ 0.005 & 0.549 $\pm$ 0.002 & 0.576 $\pm$ 0.002 \\
DialogLast+GeoAggregation & 0.636 $\pm$ 0.001 & 0.603 $\pm$ 0.001 & 0.649 $\pm$ 0.003 & 0.645 $\pm$ 0.001 & 0.648 $\pm$ 0.002 \\
DialogLast+GeoCoLES & 0.647 $\pm$ 0.002 & 0.615 $\pm$ 0.002 & 0.660 $\pm$ 0.006 & 0.650 $\pm$ 0.002 & 0.663 $\pm$ 0.003 \\
DialogLast+GeoTabGPT & 0.654 $\pm$ 0.002 & 0.631 $\pm$ 0.001 & 0.673 $\pm$ 0.005 & 0.652 $\pm$ 0.002 & 0.662 $\pm$ 0.002 \\
DialogLast+GeoTabBERT & 0.642 $\pm$ 0.003 & 0.613 $\pm$ 0.002 & 0.655 $\pm$ 0.009 & 0.646 $\pm$ 0.001 & 0.655 $\pm$ 0.004 \\
DialogLast+TrxAggregation & 0.788 $\pm$ 0.001 & 0.749 $\pm$ 0.001 & 0.826 $\pm$ 0.003 & 0.772 $\pm$ 0.001 & 0.805 $\pm$ 0.004 \\
DialogLast+TrxAggregation+GeoAggregation & 0.787 $\pm$ 0.001 & 0.746 $\pm$ 0.001 & 0.826 $\pm$ 0.004 & 0.773 $\pm$ 0.001 & 0.804 $\pm$ 0.002 \\
DialogLast+TrxCoLES & 0.776 $\pm$ 0.002 & 0.739 $\pm$ 0.003 & 0.814 $\pm$ 0.004 & 0.753 $\pm$ 0.002 & 0.797 $\pm$ 0.003 \\
DialogLast+TrxCoLES+GeoCoLES & 0.777 $\pm$ 0.002 & 0.739 $\pm$ 0.004 & 0.816 $\pm$ 0.004 & 0.755 $\pm$ 0.003 & 0.798 $\pm$ 0.002 \\
DialogLast+TrxTabGPT & 0.805 $\pm$ 0.001 & 0.775 $\pm$ 0.001 & 0.842 $\pm$ 0.002 & 0.778 $\pm$ 0.002 & 0.823 $\pm$ 0.005 \\
DialogLast+TrxTabGPT+GeoTabGPT & 0.802 $\pm$ 0.001 & 0.764 $\pm$ 0.001 & 0.845 $\pm$ 0.002 & 0.777 $\pm$ 0.001 & 0.820 $\pm$ 0.004 \\
DialogLast+TrxTabBERT & 0.764 $\pm$ 0.009 & 0.715 $\pm$ 0.019 & 0.817 $\pm$ 0.007 & 0.734 $\pm$ 0.007 & 0.789 $\pm$ 0.008 \\
DialogLast+TrxTabBERT+GeoTabBERT & 0.765 $\pm$ 0.009 & 0.714 $\pm$ 0.019 & 0.819 $\pm$ 0.006 & 0.738 $\pm$ 0.007 & 0.789 $\pm$ 0.008 \\
DialogMean & 0.604 $\pm$ 0.001 & 0.636 $\pm$ 0.002 & 0.629 $\pm$ 0.001 & 0.564 $\pm$ 0.001 & 0.587 $\pm$ 0.001 \\
DialogMean+GeoAggregation & 0.642 $\pm$ 0.001 & 0.605 $\pm$ 0.002 & 0.658 $\pm$ 0.002 & 0.650 $\pm$ 0.001 & 0.654 $\pm$ 0.001 \\
DialogMean+GeoCoLES & 0.653 $\pm$ 0.001 & 0.618 $\pm$ 0.002 & 0.670 $\pm$ 0.002 & 0.656 $\pm$ 0.001 & 0.669 $\pm$ 0.001 \\
DialogMean+GeoTabGPT & 0.661 $\pm$ 0.001 & 0.633 $\pm$ 0.001 & 0.681 $\pm$ 0.004 & 0.657 $\pm$ 0.002 & 0.671 $\pm$ 0.002 \\
DialogMean+GeoTabBERT & 0.648 $\pm$ 0.003 & 0.613 $\pm$ 0.003 & 0.666 $\pm$ 0.005 & 0.652 $\pm$ 0.002 & 0.662 $\pm$ 0.004 \\
DialogMean+TrxAggregation & 0.788 $\pm$ 0.000 & 0.749 $\pm$ 0.001 & 0.825 $\pm$ 0.003 & 0.773 $\pm$ 0.001 & 0.804 $\pm$ 0.001 \\
DialogMean+TrxAggregation+GeoAggregation & 0.787 $\pm$ 0.001 & 0.746 $\pm$ 0.001 & 0.825 $\pm$ 0.002 & 0.773 $\pm$ 0.001 & 0.804 $\pm$ 0.003 \\
DialogMean+TrxCoLES & 0.776 $\pm$ 0.002 & 0.739 $\pm$ 0.003 & 0.814 $\pm$ 0.004 & 0.753 $\pm$ 0.002 & 0.798 $\pm$ 0.003 \\
DialogMean+TrxCoLES+GeoCoLES & 0.777 $\pm$ 0.002 & 0.739 $\pm$ 0.002 & 0.815 $\pm$ 0.003 & 0.755 $\pm$ 0.003 & 0.799 $\pm$ 0.002 \\
DialogMean+TrxTabGPT & 0.805 $\pm$ 0.001 & 0.775 $\pm$ 0.001 & 0.843 $\pm$ 0.001 & 0.778 $\pm$ 0.002 & 0.824 $\pm$ 0.004 \\
DialogMean+TrxTabGPT+GeoTabGPT & 0.802 $\pm$ 0.001 & 0.764 $\pm$ 0.001 & 0.845 $\pm$ 0.002 & 0.777 $\pm$ 0.001 & 0.821 $\pm$ 0.003 \\
DialogMean+TrxTabBERT & 0.765 $\pm$ 0.009 & 0.715 $\pm$ 0.019 & 0.817 $\pm$ 0.006 & 0.735 $\pm$ 0.006 & 0.792 $\pm$ 0.007 \\
DialogMean+TrxTabBERT+GeoTabBERT & 0.766 $\pm$ 0.009 & 0.714 $\pm$ 0.019 & 0.819 $\pm$ 0.006 & 0.738 $\pm$ 0.006 & 0.792 $\pm$ 0.007 \\
GeoAggregation & 0.554 $\pm$ 0.001 & 0.540 $\pm$ 0.001 & 0.584 $\pm$ 0.002 & 0.534 $\pm$ 0.001 & 0.559 $\pm$ 0.001 \\
GeoCoLES & 0.601 $\pm$ 0.004 & 0.565 $\pm$ 0.004 & 0.668 $\pm$ 0.011 & 0.571 $\pm$ 0.003 & 0.600 $\pm$ 0.003 \\
GeoTabGPT & 0.622 $\pm$ 0.001 & 0.589 $\pm$ 0.001 & 0.700 $\pm$ 0.008 & 0.586 $\pm$ 0.003 & 0.615 $\pm$ 0.004 \\
GeoTabBERT & 0.596 $\pm$ 0.002 & 0.566 $\pm$ 0.003 & 0.663 $\pm$ 0.010 & 0.570 $\pm$ 0.003 & 0.585 $\pm$ 0.007 \\
TrxAggregation & 0.780 $\pm$ 0.005 & 0.743 $\pm$ 0.001 & 0.824 $\pm$ 0.001 & 0.762 $\pm$ 0.001 & 0.791 $\pm$ 0.017 \\
TrxAggregation+GeoAggregation & 0.779 $\pm$ 0.004 & 0.740 $\pm$ 0.001 & 0.828 $\pm$ 0.002 & 0.762 $\pm$ 0.001 & 0.787 $\pm$ 0.013 \\
TrxCoLES & 0.772 $\pm$ 0.002 & 0.734 $\pm$ 0.003 & 0.813 $\pm$ 0.004 & 0.746 $\pm$ 0.001 & 0.793 $\pm$ 0.003 \\
TrxCoLES+GeoCoLES & 0.772 $\pm$ 0.002 & 0.734 $\pm$ 0.004 & 0.814 $\pm$ 0.004 & 0.749 $\pm$ 0.002 & 0.792 $\pm$ 0.002 \\
TrxTabGPT & 0.796 $\pm$ 0.000 & 0.745 $\pm$ 0.001 & 0.837 $\pm$ 0.004 & 0.777 $\pm$ 0.001 & 0.824 $\pm$ 0.005 \\
TrxTabGPT+GeoTabGPT & 0.796 $\pm$ 0.001 & 0.751 $\pm$ 0.002 & 0.843 $\pm$ 0.003 & 0.774 $\pm$ 0.001 & 0.816 $\pm$ 0.003 \\
TrxTabBERT & 0.754 $\pm$ 0.011 & 0.707 $\pm$ 0.019 & 0.815 $\pm$ 0.006 & 0.717 $\pm$ 0.012 & 0.778 $\pm$ 0.012 \\
TrxTabBERT+GeoTabBERT & 0.756 $\pm$ 0.011 & 0.707 $\pm$ 0.019 & 0.816 $\pm$ 0.005 & 0.722 $\pm$ 0.010 & 0.778 $\pm$ 0.012 \\
\bottomrule
\end{tabular}
}
\end{table*}

\begin{table*}
 \caption{Blending results on public dataset}
\label{table:public_blending}
 \centering
 \resizebox{\textwidth}{!}{
 \begin{tabular}{l|*1{c}|*{4}{c}}
 \toprule
Methods&mean & target\_1&target\_2&target\_3& target\_4 \\
\midrule
DialogLast & 0.586 $\pm$ 0.001 & 0.602 $\pm$ 0.001 & 0.622 $\pm$ 0.004 & 0.554 $\pm$ 0.001 & 0.567 $\pm$ 0.002 \\ 
DialogLast+GeoAggregation & 0.600 $\pm$ 0.001 & 0.605 $\pm$ 0.000 & 0.648 $\pm$ 0.004 & 0.560 $\pm$ 0.001 & 0.585 $\pm$ 0.002 \\ 
DialogLast+GeoCoLES & 0.625 $\pm$ 0.003 & 0.615 $\pm$ 0.001 & 0.700 $\pm$ 0.005 & 0.577 $\pm$ 0.003 & 0.609 $\pm$ 0.003 \\ 
DialogLast+GeoTabGPT & 0.642 $\pm$ 0.002 & 0.629 $\pm$ 0.001 & 0.725 $\pm$ 0.007 & 0.589 $\pm$ 0.002 & 0.623 $\pm$ 0.002 \\ 
DialogLast+GeoTabBERT & 0.629 $\pm$ 0.001 & 0.616 $\pm$ 0.001 & 0.709 $\pm$ 0.006 & 0.577 $\pm$ 0.002 & 0.613 $\pm$ 0.001 \\ 
DialogLast+TrxAggregation & 0.732 $\pm$ 0.002 & 0.685 $\pm$ 0.001 & 0.826 $\pm$ 0.004 & 0.688 $\pm$ 0.001 & 0.731 $\pm$ 0.003 \\ 
DialogLast+TrxAggregation+GeoAggregation & 0.731 $\pm$ 0.002 & 0.681 $\pm$ 0.001 & 0.825 $\pm$ 0.005 & 0.684 $\pm$ 0.001 & 0.736 $\pm$ 0.003 \\ 
DialogLast+TrxCoLES & 0.714 $\pm$ 0.001 & 0.675 $\pm$ 0.001 & 0.806 $\pm$ 0.004 & 0.658 $\pm$ 0.002 & 0.715 $\pm$ 0.003 \\ 
DialogLast+TrxCoLES+GeoCoLES & 0.720 $\pm$ 0.001 & 0.673 $\pm$ 0.002 & 0.822 $\pm$ 0.004 & 0.657 $\pm$ 0.002 & 0.728 $\pm$ 0.002 \\ 
DialogLast+TrxTabGPT & 0.743 $\pm$ 0.002 & 0.677 $\pm$ 0.001 & 0.834 $\pm$ 0.004 & 0.697 $\pm$ 0.001 & 0.766 $\pm$ 0.003 \\ 
DialogLast+TrxTabGPT+GeoTabGPT & 0.753 $\pm$ 0.001 & 0.684 $\pm$ 0.001 & 0.848 $\pm$ 0.004 & 0.700 $\pm$ 0.001 & 0.779 $\pm$ 0.003 \\ 
DialogLast+TrxTabBERT & 0.710 $\pm$ 0.005 & 0.669 $\pm$ 0.003 & 0.814 $\pm$ 0.006 & 0.640 $\pm$ 0.008 & 0.715 $\pm$ 0.007 \\ 
DialogLast+TrxTabBERT+GeoTabBERT & 0.718 $\pm$ 0.004 & 0.670 $\pm$ 0.003 & 0.828 $\pm$ 0.005 & 0.641 $\pm$ 0.008 & 0.733 $\pm$ 0.006 \\ 
DialogMean & 0.595 $\pm$ 0.002 & 0.600 $\pm$ 0.001 & 0.633 $\pm$ 0.006 & 0.566 $\pm$ 0.001 & 0.580 $\pm$ 0.002 \\ 
DialogMean+GeoAggregation & 0.607 $\pm$ 0.001 & 0.604 $\pm$ 0.000 & 0.657 $\pm$ 0.004 & 0.572 $\pm$ 0.000 & 0.596 $\pm$ 0.002 \\ 
DialogMean+GeoCoLES & 0.630 $\pm$ 0.002 & 0.615 $\pm$ 0.001 & 0.703 $\pm$ 0.006 & 0.586 $\pm$ 0.002 & 0.618 $\pm$ 0.003 \\ 
DialogMean+GeoTabGPT & 0.645 $\pm$ 0.002 & 0.629 $\pm$ 0.001 & 0.727 $\pm$ 0.010 & 0.596 $\pm$ 0.001 & 0.630 $\pm$ 0.001 \\ 
DialogMean+GeoTabBERT & 0.634 $\pm$ 0.002 & 0.616 $\pm$ 0.002 & 0.713 $\pm$ 0.006 & 0.586 $\pm$ 0.002 & 0.621 $\pm$ 0.002 \\ 
DialogMean+TrxAggregation & 0.732 $\pm$ 0.001 & 0.684 $\pm$ 0.001 & 0.824 $\pm$ 0.001 & 0.688 $\pm$ 0.001 & 0.732 $\pm$ 0.002 \\ 
DialogMean+TrxAggregation+GeoAggregation & 0.731 $\pm$ 0.001 & 0.679 $\pm$ 0.000 & 0.822 $\pm$ 0.002 & 0.684 $\pm$ 0.001 & 0.737 $\pm$ 0.002 \\ 
DialogMean+TrxCoLES & 0.713 $\pm$ 0.001 & 0.674 $\pm$ 0.001 & 0.804 $\pm$ 0.002 & 0.659 $\pm$ 0.002 & 0.717 $\pm$ 0.003 \\ 
DialogMean+TrxCoLES+GeoCoLES & 0.719 $\pm$ 0.001 & 0.672 $\pm$ 0.002 & 0.819 $\pm$ 0.003 & 0.658 $\pm$ 0.002 & 0.729 $\pm$ 0.002 \\ 
DialogMean+TrxTabGPT & 0.742 $\pm$ 0.001 & 0.675 $\pm$ 0.001 & 0.829 $\pm$ 0.004 & 0.697 $\pm$ 0.001 & 0.766 $\pm$ 0.003 \\ 
DialogMean+TrxTabGPT+GeoTabGPT & 0.751 $\pm$ 0.001 & 0.682 $\pm$ 0.001 & 0.845 $\pm$ 0.004 & 0.700 $\pm$ 0.001 & 0.779 $\pm$ 0.004 \\ 
DialogMean+TrxTabBERT & 0.709 $\pm$ 0.004 & 0.668 $\pm$ 0.003 & 0.812 $\pm$ 0.008 & 0.642 $\pm$ 0.007 & 0.717 $\pm$ 0.006 \\ 
DialogMean+TrxTabBERT+GeoTabBERT & 0.718 $\pm$ 0.003 & 0.668 $\pm$ 0.003 & 0.826 $\pm$ 0.006 & 0.643 $\pm$ 0.007 & 0.734 $\pm$ 0.006 \\
GeoAggregation & 0.555 $\pm$ 0.001 & 0.539 $\pm$ 0.000 & 0.590 $\pm$ 0.002 & 0.533 $\pm$ 0.001 & 0.560 $\pm$ 0.001 \\ 
GeoCoLES & 0.598 $\pm$ 0.004 & 0.568 $\pm$ 0.003 & 0.663 $\pm$ 0.005 & 0.568 $\pm$ 0.007 & 0.593 $\pm$ 0.005 \\ 
GeoTabGPT & 0.621 $\pm$ 0.003 & 0.589 $\pm$ 0.002 & 0.696 $\pm$ 0.010 & 0.586 $\pm$ 0.002 & 0.614 $\pm$ 0.002 \\ 
GeoTabBERT & 0.603 $\pm$ 0.002 & 0.573 $\pm$ 0.003 & 0.672 $\pm$ 0.007 & 0.570 $\pm$ 0.004 & 0.598 $\pm$ 0.004 \\ 
TrxAggregation & 0.788 $\pm$ 0.001 & 0.743 $\pm$ 0.003 & 0.831 $\pm$ 0.002 & 0.777 $\pm$ 0.001 & 0.800 $\pm$ 0.002 \\ 
TrxAggregation+GeoAggregation & 0.778 $\pm$ 0.002 & 0.733 $\pm$ 0.002 & 0.822 $\pm$ 0.004 & 0.767 $\pm$ 0.001 & 0.790 $\pm$ 0.002 \\ 
TrxCoLES & 0.774 $\pm$ 0.002 & 0.734 $\pm$ 0.002 & 0.812 $\pm$ 0.004 & 0.759 $\pm$ 0.002 & 0.790 $\pm$ 0.003 \\ 
TrxCoLES+GeoCoLES & 0.775 $\pm$ 0.001 & 0.730 $\pm$ 0.002 & 0.827 $\pm$ 0.004 & 0.751 $\pm$ 0.003 & 0.792 $\pm$ 0.002 \\ 
TrxTabGPT & 0.802 $\pm$ 0.001 & 0.751 $\pm$ 0.001 & 0.844 $\pm$ 0.002 & 0.788 $\pm$ 0.002 & 0.826 $\pm$ 0.003 \\ 
TrxTabGPT+GeoTabGPT & 0.804 $\pm$ 0.001 & 0.748 $\pm$ 0.001 & 0.854 $\pm$ 0.002 & 0.784 $\pm$ 0.001 & 0.829 $\pm$ 0.003 \\ 
TrxTabBERT & 0.762 $\pm$ 0.004 & 0.717 $\pm$ 0.006 & 0.819 $\pm$ 0.004 & 0.734 $\pm$ 0.006 & 0.778 $\pm$ 0.006 \\ 
TrxTabBERT+GeoTabBERT & 0.766 $\pm$ 0.004 & 0.717 $\pm$ 0.006 & 0.831 $\pm$ 0.005 & 0.729 $\pm$ 0.007 & 0.786 $\pm$ 0.005 \\ 
\bottomrule
\end{tabular}
}
\end{table*}

\begin{table*}
 \caption{Late Fusion results on public dataset}
\label{table:public_fusion}
 \centering
 \resizebox{\textwidth}{!}{
 \begin{tabular}{l|*1{c}|*{4}{c}}
 \toprule
Methods&mean & target\_1&target\_2&target\_3& target\_4 \\
\midrule
DialogLast & 0.586 $\pm$ 0.001 & 0.602 $\pm$ 0.001 & 0.622 $\pm$ 0.004 & 0.554 $\pm$ 0.001 & 0.567 $\pm$ 0.002 \\
DialogLast+GeoAggregation & 0.646 $\pm$ 0.001 & 0.614 $\pm$ 0.001 & 0.659 $\pm$ 0.003 & 0.654 $\pm$ 0.001 & 0.657 $\pm$ 0.001 \\
DialogLast+GeoCoLES & 0.660 $\pm$ 0.001 & 0.633 $\pm$ 0.002 & 0.675 $\pm$ 0.004 & 0.661 $\pm$ 0.001 & 0.671 $\pm$ 0.003 \\
DialogLast+GeoTabGPT & 0.668 $\pm$ 0.001 & 0.645 $\pm$ 0.001 & 0.690 $\pm$ 0.004 & 0.662 $\pm$ 0.001 & 0.674 $\pm$ 0.002 \\
DialogLast+GeoTabBERT & 0.662 $\pm$ 0.001 & 0.633 $\pm$ 0.002 & 0.680 $\pm$ 0.003 & 0.662 $\pm$ 0.001 & 0.675 $\pm$ 0.002 \\
DialogLast+TrxAggregation & 0.792 $\pm$ 0.002 & 0.752 $\pm$ 0.001 & 0.829 $\pm$ 0.001 & 0.780 $\pm$ 0.001 & 0.805 $\pm$ 0.006 \\
DialogLast+TrxAggregation+GeoAggregation & 0.791 $\pm$ 0.001 & 0.750 $\pm$ 0.001 & 0.829 $\pm$ 0.005 & 0.782 $\pm$ 0.001 & 0.803 $\pm$ 0.001 \\
DialogLast+TrxCoLES & 0.782 $\pm$ 0.001 & 0.746 $\pm$ 0.003 & 0.814 $\pm$ 0.003 & 0.765 $\pm$ 0.002 & 0.802 $\pm$ 0.003 \\
DialogLast+TrxCoLES+GeoCoLES & 0.783 $\pm$ 0.001 & 0.745 $\pm$ 0.004 & 0.819 $\pm$ 0.003 & 0.767 $\pm$ 0.001 & 0.803 $\pm$ 0.002 \\
DialogLast+TrxTabGPT & 0.810 $\pm$ 0.001 & 0.779 $\pm$ 0.001 & 0.846 $\pm$ 0.003 & 0.789 $\pm$ 0.002 & 0.827 $\pm$ 0.004 \\
DialogLast+TrxTabGPT+GeoTabGPT & 0.808 $\pm$ 0.001 & 0.770 $\pm$ 0.001 & 0.849 $\pm$ 0.003 & 0.790 $\pm$ 0.001 & 0.824 $\pm$ 0.004 \\
DialogLast+TrxTabBERT & 0.773 $\pm$ 0.002 & 0.730 $\pm$ 0.006 & 0.822 $\pm$ 0.005 & 0.749 $\pm$ 0.004 & 0.792 $\pm$ 0.003 \\
DialogLast+TrxTabBERT+GeoTabBERT & 0.776 $\pm$ 0.003 & 0.729 $\pm$ 0.006 & 0.827 $\pm$ 0.004 & 0.752 $\pm$ 0.004 & 0.794 $\pm$ 0.003 \\
DialogMean & 0.595 $\pm$ 0.002 & 0.600 $\pm$ 0.001 & 0.633 $\pm$ 0.006 & 0.566 $\pm$ 0.001 & 0.580 $\pm$ 0.002 \\
DialogMean+GeoAggregation & 0.649 $\pm$ 0.001 & 0.614 $\pm$ 0.001 & 0.665 $\pm$ 0.002 & 0.656 $\pm$ 0.001 & 0.662 $\pm$ 0.001 \\
DialogMean+GeoCoLES & 0.663 $\pm$ 0.001 & 0.632 $\pm$ 0.002 & 0.680 $\pm$ 0.004 & 0.663 $\pm$ 0.000 & 0.675 $\pm$ 0.002 \\
DialogMean+GeoTabGPT & 0.670 $\pm$ 0.001 & 0.645 $\pm$ 0.000 & 0.694 $\pm$ 0.004 & 0.664 $\pm$ 0.001 & 0.678 $\pm$ 0.001 \\
DialogMean+GeoTabBERT & 0.664 $\pm$ 0.001 & 0.633 $\pm$ 0.001 & 0.682 $\pm$ 0.002 & 0.664 $\pm$ 0.001 & 0.678 $\pm$ 0.001 \\
DialogMean+TrxAggregation & 0.792 $\pm$ 0.002 & 0.752 $\pm$ 0.002 & 0.828 $\pm$ 0.002 & 0.781 $\pm$ 0.001 & 0.807 $\pm$ 0.006 \\
DialogMean+TrxAggregation+GeoAggregation & 0.792 $\pm$ 0.002 & 0.750 $\pm$ 0.002 & 0.829 $\pm$ 0.002 & 0.782 $\pm$ 0.001 & 0.807 $\pm$ 0.005 \\
DialogMean+TrxCoLES & 0.781 $\pm$ 0.001 & 0.745 $\pm$ 0.003 & 0.814 $\pm$ 0.002 & 0.765 $\pm$ 0.002 & 0.802 $\pm$ 0.003 \\
DialogMean+TrxCoLES+GeoCoLES & 0.783 $\pm$ 0.001 & 0.744 $\pm$ 0.004 & 0.819 $\pm$ 0.003 & 0.767 $\pm$ 0.002 & 0.802 $\pm$ 0.002 \\
DialogMean+TrxTabGPT & 0.810 $\pm$ 0.002 & 0.779 $\pm$ 0.001 & 0.847 $\pm$ 0.004 & 0.789 $\pm$ 0.001 & 0.828 $\pm$ 0.003 \\
DialogMean+TrxTabGPT+GeoTabGPT & 0.808 $\pm$ 0.001 & 0.770 $\pm$ 0.001 & 0.848 $\pm$ 0.003 & 0.790 $\pm$ 0.001 & 0.826 $\pm$ 0.003 \\
DialogMean+TrxTabBERT & 0.773 $\pm$ 0.003 & 0.730 $\pm$ 0.006 & 0.822 $\pm$ 0.004 & 0.749 $\pm$ 0.004 & 0.791 $\pm$ 0.003 \\
DialogMean+TrxTabBERT+GeoTabBERT & 0.775 $\pm$ 0.003 & 0.728 $\pm$ 0.005 & 0.827 $\pm$ 0.003 & 0.752 $\pm$ 0.004 & 0.794 $\pm$ 0.004 \\
GeoAggregation & 0.555 $\pm$ 0.001 & 0.539 $\pm$ 0.000 & 0.590 $\pm$ 0.002 & 0.533 $\pm$ 0.001 & 0.560 $\pm$ 0.001 \\
GeoCoLES & 0.598 $\pm$ 0.004 & 0.568 $\pm$ 0.003 & 0.663 $\pm$ 0.005 & 0.568 $\pm$ 0.007 & 0.593 $\pm$ 0.005 \\
GeoTabGPT & 0.621 $\pm$ 0.003 & 0.589 $\pm$ 0.002 & 0.696 $\pm$ 0.010 & 0.586 $\pm$ 0.002 & 0.614 $\pm$ 0.002 \\
GeoTabBERT & 0.603 $\pm$ 0.002 & 0.573 $\pm$ 0.003 & 0.672 $\pm$ 0.007 & 0.570 $\pm$ 0.004 & 0.598 $\pm$ 0.004 \\
TrxAggregation & 0.783 $\pm$ 0.002 & 0.741 $\pm$ 0.003 & 0.828 $\pm$ 0.003 & 0.770 $\pm$ 0.004 & 0.792 $\pm$ 0.007 \\
TrxAggregation+GeoAggregation & 0.783 $\pm$ 0.002 & 0.740 $\pm$ 0.002 & 0.829 $\pm$ 0.003 & 0.771 $\pm$ 0.001 & 0.792 $\pm$ 0.011 \\
TrxCoLES & 0.773 $\pm$ 0.002 & 0.734 $\pm$ 0.002 & 0.812 $\pm$ 0.004 & 0.758 $\pm$ 0.002 & 0.790 $\pm$ 0.003 \\
TrxCoLES+GeoCoLES & 0.775 $\pm$ 0.002 & 0.734 $\pm$ 0.002 & 0.815 $\pm$ 0.004 & 0.760 $\pm$ 0.002 & 0.789 $\pm$ 0.003 \\
TrxTabGPT & 0.802 $\pm$ 0.001 & 0.751 $\pm$ 0.001 & 0.844 $\pm$ 0.002 & 0.787 $\pm$ 0.001 & 0.825 $\pm$ 0.003 \\
TrxTabGPT+GeoTabGPT & 0.800 $\pm$ 0.001 & 0.752 $\pm$ 0.001 & 0.846 $\pm$ 0.005 & 0.785 $\pm$ 0.002 & 0.817 $\pm$ 0.006 \\
TrxTabBERT & 0.762 $\pm$ 0.004 & 0.717 $\pm$ 0.006 & 0.819 $\pm$ 0.004 & 0.734 $\pm$ 0.006 & 0.777 $\pm$ 0.006 \\
TrxTabBERT+GeoTabBERT & 0.764 $\pm$ 0.004 & 0.716 $\pm$ 0.006 & 0.823 $\pm$ 0.004 & 0.737 $\pm$ 0.006 & 0.780 $\pm$ 0.005 \\
\bottomrule
\end{tabular}
}
\end{table*}

\end{document}